%% file: bali.tex
\documentclass{article}

\usepackage[nonatbib,preprint]{neurips_2024}

\usepackage[utf8]{inputenc} 
\usepackage[T1]{fontenc}    
\usepackage[hidelinks]{hyperref}       
\usepackage{url}            
\usepackage{booktabs}       
\usepackage{amsfonts}       
\usepackage{nicefrac}       
\usepackage{microtype}      
\usepackage{xcolor}         

\usepackage{graphicx}
\usepackage{subfigure}

\usepackage{amsmath}
\usepackage{amssymb}
\usepackage{mathtools}
\usepackage{amsthm}
\usepackage{nicefrac}
\usepackage[utf8]{inputenc} 

\usepackage[capitalize,noabbrev]{cleveref}

\usepackage{enumitem}
\setlist{noitemsep, topsep=0pt, leftmargin=*}
\usepackage{xfrac}

\usepackage{booktabs}
\usepackage{adjustbox}
\usepackage{rotating}

\theoremstyle{plain}
\newtheorem{property}{Property}[section]

\theoremstyle{definition}

\theoremstyle{remark}

\usepackage[textsize=tiny]{todonotes}

\usepackage{letltxmacro}
\LetLtxMacro{\originaleqref}{\eqref}
\renewcommand{\eqref}{\normalfont{Eq.}~\originaleqref}

\newcommand{\by}{\mathbf{y}}
\newcommand{\bx}{\mathbf{x}}
\newcommand{\bz}{\mathbf{z}}
\newcommand{\ba}{\mathbf{a}}

\newcommand{\bm}{\mathbf{m}}

\newcommand{\rmu}{\mathrm{u}}

\newcommand{\bY}{\mathbf{Y}}
\newcommand{\bX}{\mathbf{X}}
\newcommand{\bV}{\mathbf{V}}
\newcommand{\bU}{\mathbf{U}}
\newcommand{\bR}{\mathbf{R}}
\newcommand{\bH}{\mathbf{H}}
\newcommand{\bA}{\mathbf{A}}
\newcommand{\bB}{\mathbf{B}}
\newcommand{\bC}{\mathbf{C}}
\newcommand{\bD}{\mathbf{D}}
\newcommand{\bE}{\mathbf{E}}

\newcommand{\bSigma}{\boldsymbol{\Sigma}}

\newcommand\Eta{\mathbf{H}}
\newcommand\bM{\mathbf{M}}

\newcommand{\bPsi}{\boldsymbol{\Psi}}
\newcommand{\btheta}{\boldsymbol{\theta}}

\newcommand{\weights}{\mathbf{w}}
\newcommand{\Weights}{\mathbf{W}}
\newcommand{\Dataset}{\mathcal{D}}

\newcommand{\kron}{\otimes}

\usepackage{algorithm}
\usepackage{algorithmic}
\usepackage{wrapfig}

\title{BALI: Learning Neural Networks via \\ Bayesian Layerwise Inference}

\author{%
	Richard Kurle\thanks{Correspondence to richard.kurle@tum.de. Work done while at Hasso Plattner Institute.} \\
	NXAI GmbH \\
	Linz, Austria 
	\And
	Alexej Klushyn \\
	Airbus Central Research \& Technology \\
	Munich, Germany 
	\And
	Ralf Herbrich \\
	Hasso Plattner Institute \\
	Potsdam, Germany 
}

\begin{document}
	\maketitle

\input{sec/abstract}
\input{sec/introduction}
\input{sec/background}
\input{sec/method}
\input{sec/related_work}
\input{sec/experiments}
\input{sec/conclusion}

\section*{Acknowledgements}
This research was partially funded by Airbus Central Research \& Technology, AI Research.


\input{bali.bbl}
\newpage
\appendix

\input{sec/appendix/kronecker_properties}
\input{sec/appendix/posterior_parameters}
\input{sec/appendix/natural_parameters}
\input{sec/appendix/complexity}

\input{sec/appendix/algorithm_comparison}
\input{sec/appendix/mean_updates}
\input{sec/appendix/experiment_details}

\end{document}

%% file: sec/abstract.tex
\begin{abstract}
We introduce a new method for learning Bayesian neural networks, treating them as a stack of multivariate Bayesian linear regression models. The main idea is to infer the layerwise posterior exactly if we know the target outputs of each layer. We define these pseudo-targets as the layer outputs from the forward pass, updated by the backpropagated gradients of the objective function. The resulting layerwise posterior is a matrix-normal distribution with a Kronecker-factorized covariance matrix, which can be efficiently inverted. Our method extends to the stochastic mini-batch setting using an exponential moving average over natural-parameter terms, thus gradually forgetting older data. The method converges in few iterations and performs as well as or better than leading Bayesian neural network methods on various regression, classification, and out-of-distribution detection benchmarks. 
\end{abstract}

%% file: sec/introduction.tex
\section{Introduction}
Modeling and inferring the weights of neural networks (NN) in a probabilistic manner promises many advantages compared to deterministic point estimates: 
quantifying the epistemic uncertainty leads to better calibration and generalisation to unseen data \cite{Wilson2022Bayesian} and the probabilistic framework naturally extends to important practical applications such as continual \cite{Ritter2018LA, Nguyen2018VCL, Kurle2020LLL} and active learning \cite{Gal2017Active, Bickford2023prediction}. 

Variational inference methods \cite{Blundell2015a, Hinton1993a} are particularly appealing due to their computational scalability to larger models and dataset sizes. 
However, these methods often yield poor fits, especially for large model sizes \cite{Ghosh2018}, which has so far impeded the use of these methods in practical applications.
This issue has recently been attributed to invariances in the parametrisation of the likelihood of deep NNs, which lead to an additional gap between the evidence lower bound objective and the log-marginal likelihood \cite{Kurle2022VBNN}.
For instance, the invariance w.r.t.\ all possible node permutations within each hidden layer gives rise to an intractable amount of modes that is factorial in the number of neurons per hidden layer. 
As a consequence, Gaussian (uni-modal) posterior approximations prefer solutions for which these modes overlap, such as when the posterior reverts to the zero-centered prior.

Notably, certain inference methods implicitly circumvent this problem through a \emph{local linear} approximation of the NN. 
For instance, practical Laplace approximations \cite{MacKay1992a, Daxberger2021Laplace} use a \emph{local} Gaussian approximation around a point estimate with the Generalized Gauss-Newton (GGN) matrix to approximate the Hessian. 
Similarly, noisy K-FAC \cite{Zhang18NoisyKFAC} uses the GGN to approximate the Fisher information matrix at a weight sample from the current posterior. 
And in \cite{Ober2021GlobalInducingPoint}, a small number of learned inducing points is propagated through the NN, defining a local posterior approximation as the solution to the Bayesian linear regression problem using the inducing points as pseudo-inputs and pseudo-targets.

Building on these insights, we propose \emph{Bayesian Layerwise Inference} (BALI), a new method for learning Bayesian neural networks (BNN). 
The central idea is that, for every layer, the local posterior can be inferred exactly and analytically, given the features extracted \emph{forward} from the preceding layers and pseudo-targets projected \emph{backward} from the subsequent layers. 
The posterior approximation therefore consists of layerwise Gaussian distributions given by the solution to a multivariate Bayesian linear regression problem. 
By treating BNNs as a stack of linear models, this approach effectively breaks the troubling node-permutation invariance, because each layer optimises a locally convex objective function which depends on other layers only through the propagated inputs and targets. 
In this approximate inference framework, the major challenges involve computing the pseudo-targets and scaling the approximate inference method to larger models and datasets via mini-batching. 
Addressing these problems, our contributions are as follows:
\begin{itemize}
\item we propose BALI, a novel approximate inference method for BNNs, leveraging locally exact inference in layerwise linear regression models (Sec.~\ref{sec:method:inference});
\item we then extend this method to the mini-batch setting by using an exponential moving average estimate of natural parameter terms stemming from the likelihood (Sec.~\ref{sec:method:adaptation});
\item building on gradient back-propagation, we compute the regression pseudo-targets for each layer as the gradient-updated layer outputs (Sec.~\ref{sec:method:targets}). 
\item we evaluate BALI on regression, classification and out-of-distribution detection tasks; it achieves similar or better performance compared to strong BNN baselines, while converging quickly (Sec.~\ref{sec:experiments}). 
\end{itemize}

%% file: sec/background.tex
\section{Background}
\subsection{Gaussian linear models}\label{sec:background:univariate_linear_regression}
Given a dataset of size $N$ with inputs $\bX \in \mathbb{R}^{N \times D_{\bx}}$ and targets $\by \in \mathbb{R}^{N}$,
linear models predict the targets via a linear transformation $\bz=\bX \weights$ of the inputs with a weight vector $\weights \in \mathbb{R}^{D_{\bx}}$. 
Gaussian linear models further assume that the weights and prediction errors are normal distributed, which is expressed by the following likelihood and prior:
\begin{align}
	p(\by | \bX, \weights) &= \mathcal{N}(\bX \weights, \bSigma), \label{eq:univariate_likelihood} 
		~~~~~ p(\weights) = \mathcal{N}(\bm_0,  \bV_0).
\end{align}
The prior mean vector $\bm_0 \in \mathbb{R}^{D_{\bx}}$ and covariance matrix ${\bV_0 \in \mathbb{R}^{D_{\bx} \times D_{\bx}}}$ as well as the noise covariance $\bSigma \in \mathbb{R}^{N \times N}$ are hyperparameters of the model.
Optional bias terms can be incorporated by concatenating a vector $\mathbf{1}$ of size $N$ to $\bX$. 

Since the likelihood is a Gaussian function of the weights, the \emph{posterior} distribution
$
    p(\weights | \bX, \by) \propto p(\weights)\, p(\by | \bX, \weights)
$
is also Gaussian, i.e. $p(\weights | \bX, \by) = \mathcal{N}(\bm, \bV)$. 
It is easiest to express the product of the Gaussian prior and likelihood as the sum of their respective \emph{natural parameters} \cite{MurphyProbabilisticPerspective}
\begin{subequations}\label{eq:GLM:posterior_natural_params}
\begin{align}
	\bV^{-1} &= \bV_0^{-1} + \bV_{1:N}^{-1} = \bV_0^{-1} + \bX^\top \bSigma^{-1} \bX, \\
	\boldsymbol{\eta} &= \boldsymbol{\eta}_0 + \boldsymbol{\eta}_{1:N} = \bV_0^{-1}\bm_0 + \bX^\top \bSigma^{-1} \by.
\end{align}
\end{subequations}
The vector $\boldsymbol{\eta} = \bV^{-1} \bm$ is referred to as the precision-mean, $\bV^{-1}$ is the precision matrix. 
Subscripts indicate the prior and likelihood terms from $N$ data points.
The covariance $\bV$ is then computed via matrix inversion and the mean is $\bm = \bV \boldsymbol{\eta}$.

The \emph{posterior-predictive} distribution for a test data point ${(\bx_\ast, y_\ast})$ is also Gaussian:
\begin{align}
p(y_\ast| \bx_\ast, \by, \bX) 
&= \int\! p(y_\ast | \bx_\ast, \weights)\, p(\weights | \bX, \by)\, \mathrm{d}\weights 
~= \mathcal{N}(\bx_\ast^{\top} \bm,\: \bx_\ast^{\top} \bV \bx_\ast\! + \sigma^2_\ast),
\end{align}
where $\sigma^2_\ast$ is the noise variance of $p(y_\ast | \bx_\ast, \weights)$, often assumed identical for all targets (homoscedastic).

\subsection{Multivariate Bayesian linear regression (known $\bSigma$)}\label{sec:background:MVBLR}
Multivariate Bayesian linear regression assumes $N$ correlated targets $\by_n \in \mathbb{R}^{D_\by}$ and models them as
\begin{align}\label{eq:multivariate_single_obs}
\by_n = \Weights^\top \bx_n + \boldsymbol{\epsilon}_n, ~~~~~ \boldsymbol{\epsilon}_n \sim \mathcal{N}(\mathbf{0}, \bSigma),
\end{align}
where $\bz_n := \Weights^\top \bx_n$ is the prediction of the $n$-th target via a linear transformation of the inputs with a weight matrix $\Weights \in \mathbb{R}^{D_\bx \times D_\by}$, and the noise covariance matrix $\bSigma \in \mathbb{R}^{D_\by \times D_\by}$ is a hyperparameter. 

To compute the posterior of this model, it is helpful to make the connection to the univariate linear model by vectorising the weights ${\weights \coloneqq \mathrm{vec}(\Weights)}$, concatenating columns of the weight matrix, each corresponding to weights connected to the same output node.
This results in the per-data likelihood
\begin{align}
    p(\by_n | \bx_n, \weights) &= \mathcal{N}\big((\mathbf{I}_{D_\by} \kron \bx_n^\top)  \weights, \bSigma\big).
\end{align}
It can be seen that the vectorisation of the linear transformation in \eqref{eq:multivariate_single_obs} leads to a Kronecker product (cf.~App.~\ref{app:kron}, \eqref{eq:kronecker:definition}) between an input vector $\bx_n$ and the identity matrix $\mathbf{I}_{D_\by}$ of size $D_\by$, resulting in a matrix $\bar{\bX}_n := \mathbf{I}_{D_\by} \kron \bx_n^\top$ with $\bar{\bX}_n \in \mathbb{R}^{D_\by \times D_\by \cdot D_\bx}$. 
This per-data likelihood with $D_\by$-dimensional targets now has the same form as the univariate likelihood in \eqref{eq:univariate_likelihood} with $D_\by$ data points. 
As a result, each likelihood term is a Gaussian function of the weights and has natural parameters $\bV^{-1}_n = \bar{\bX}_n^\top \bSigma^{-1} \bar{\bX}_n$ and $\boldsymbol{\eta}_{n} = \bar{\bX}_n^\top \bSigma^{-1} \by_n$, analogous to \eqref{eq:GLM:posterior_natural_params}.
Fortunately, we can simplify these terms due to the Kronecker-product structure of $\bar{\bX}_n$ (cf.~App.~\ref{sec:appendix:MVBLR}):
\begin{subequations}
\begin{align}
    \bV_n^{-1} 
    &= \bSigma^{-1} \kron \bx_n \bx_n^\top
    = \bSigma^{-1} \kron \bR_n^{-1}, \\
    \boldsymbol{\eta}_{n} 
    &= \mathrm{vec} \left( \bx_n \by_n^\top \bSigma^{-1} \right) 
    = \mathrm{vec} \left( \Eta_n \right).
\end{align}
\end{subequations}
Here and in the following, we denote the second factor of the precision matrix by $\bR_n \in \mathbb{R}^{D_\bx \times D_\bx}$ and the precision-mean in matrix shape as $\Eta_n \in \mathbb{R}^{D_\bx \times D_\by}$.

Taking the product of the $N$ per-data likelihood terms, $p(\bY | \bX, \weights) = \prod_{n=1}^{N} p(\by_n | \bx_n, \weights)$, where $\bX \in \mathbb{R}^{N \times D_\bx}$ and $\bY \in \mathbb{R}^{N \times D_\by}$, the natural parameters of the likelihood of the entire dataset are
\begin{subequations}\label{eq:MVBLR:likelihood:natural_params}
\begin{align}
\bV^{-1}_{1:N} &= \bSigma^{-1} \kron (\bX^\top \bX) = \bSigma^{-1} \kron \bR^{-1}_{1:N}, \label{eq:MVBLR:likelihood:precision}\\
\Eta_{1:N} &= \bX^\top \bY \bSigma^{-1}. \label{eq:MVBLR:likelihood:precision-mean}
\end{align}
\end{subequations}
The conjugate prior for this likelihood is a matrix-normal distribution, which is a generalisation of the multivariate normal distribution to matrix-valued random variables via
\begin{equation}\label{eq:MVBLR:prior}
 p(\Weights | \bSigma) = \mathcal{MN}(\bM_0, \bR_0, \bSigma),
 ~~~~~
 p(\weights | \bSigma) = \mathcal{N}(\bm_0, \bSigma \kron \bR_0),
\end{equation}
such that the distribution over the vectorised weights $\weights$ has vectorised means $\bm_0 = \mathrm{vec}(\bM_0)$ and the covariance $\bV_0 = \bSigma \kron \bR_0$ has a Kronecker-product structure. 
Note that the prior is a conditional distribution that takes the noise covariance $\bSigma$ as one scale matrix and has a second scale matrix $\bR_0 \in \mathbb{R}^{D_\bx \times D_\bx}$.
Due to the conjugacy, the posterior is also a matrix-normal distribution
\begin{align}\label{eq:MVBLR:posterior}
    p(\Weights | \bX, \bY, \bSigma) &= \mathcal{MN}(\bM, \bR, \bSigma)
\end{align}
for which the natural parameters are given by the sum of the respective prior and likelihood terms:
\begin{subequations}\label{eq:MVBLR:natural_params}
\begin{align}
&\bV^{-1} = \bV_0^{-1} + \bV_{1:N}^{-1} = \bSigma^{-1} \kron (\bR_0^{-1} + \bX^\top \bX), \label{eq:MVBLR:precision}\\
&\Eta = \Eta_0 + \Eta_{1:N} = \bR_0^{-1} \bM_0 \bSigma^{-1} + \bX^\top \bY \bSigma^{-1}. \label{eq:MVBLR:precision-mean}
\end{align}
\end{subequations}
The distribution parameters can then be computed by (cf.~App.~\ref{sec:appendix:MVBLR})
\begin{subequations}\label{eq:MVBLR:distribution_params}
\begin{align}
    \bV &= \bSigma \kron \bR = \bSigma \kron (\bR_0^{-1} + \bX^\top \bX)^{-1}, \label{eq:MVBLR:covariance}\\
    \bM &= \bR \bH \bSigma = \bR\, (\bR_0^{-1} \bM_0 + \bX^\top \bY ).\label{eq:MVBLR:mean}
\end{align}
\end{subequations}
The inversion of the precision $\bR^{-1}$ only requires inverting the two Kronecker-product terms $\bSigma^{-1}$ and $\bR^{-1}$. 
Note also that the posterior mean is independent of the noise covariance $\bSigma$, because we defined the prior covariance proportional to $\bSigma$ via the Kronecker product in \eqref{eq:MVBLR:prior}.

\subsection{Bayesian neural networks}\label{sec:BNN}
Deep NNs are layered models that progressively transform the inputs by a sequence of linear transformations and non-linear activation functions. 
Modern architectures also include some form of normalisation between the layers (e.g.~\cite{Ioffe2015,Ba16}) to prevent the features norms from exploding or vanishing. 
Here we consider a simpler setup, dividing the features by the square-root of its dimension.
This is similar to scaling the weights in standard initialisation methods but allows us to use priors with the same scale in each layer.
Thus, a NN with $L$ layers and $D_l$ hidden nodes in layer $l$ computes
\begin{align}\label{eq:neural_network_function}
	\bx_{n,l} &= \frac{1}{\sqrt{D_{l-1} + 1}} \cdot \begin{pmatrix}  h(\bz_{n,l-1}) \\ 1  \end{pmatrix},
	~~~~~
	\bz_{n,l} = \Weights^\top_l \bx_{n,l}, 
\end{align}
where we include the bias vector in the weight matrix $\Weights_l \in \mathbb{R}^{(D_{l-1} + 1) \times D_l}$ and concatenate constant values $1$ to the features $h(\bz_{n,l-1})$ as in Sec.~\ref{sec:background:univariate_linear_regression} and Sec.~\ref{sec:background:MVBLR}.  
Here we take the pre-activation view, whereby each layer first applies the non-linear transfer function $h$ and then computes the linear transformation. 
The first layer takes the actual data (or features) as input, i.e.\ $\bx_{n,1}^\top = \left( \bx_n^\top, 1 \right)$.

A BNN further assumes a prior $p(\weights)$ over the weights. 
We denote the vector of all weights by the concatenation $\weights^\top\! \coloneqq \left(\weights_1^\top, \, \dots, \, \weights_L^\top \right)$, with $\weights_l = \mathrm{vec}(\Weights_l)$, analogous to Sec.~\ref{sec:background:MVBLR}. 
The likelihood $p(\bY | \bX, \weights) = \prod_{n=1}^{N} p(\by_n | \bx_n, \weights)$ is parametrised by the outputs of the last layer, i.e.\ $p(\by_n | \bx_n, \weights) = p(\by_n | \bz_{n, L})$.
We then combine the prior and likelihood via Bayes' rule,
$
p(\weights | \bY, \bX) = \frac{p(\bY | \bX, \weights)\, p(\weights)}{p(\bY | \bX)}.
$
Since $p(\bY | \bX) = \int\! p(\bY | \bX, \weights)\, p(\weights)\, \mathrm{d}\weights$ is intractable and $p(\bY | \bX, \weights)\, p(\weights)$ has no simple analytical form, the posterior must be approximated. 
Approximate inference in BNNs is complicated by the high multi-modality of the posterior, especially due to invariances in their likelihood function. 
For instance, any permutation of the hidden nodes---which can be expressed by the simultaneous permutation of the in-going and out-going weight matrix---leads to the same likelihood. 
As a consequence, when using a standard normal prior, the posterior has $\prod_{l=1}^{L-1} D_l !$ equivalent modes and is thus intractable to compute \cite{Kurle2021BDL}. 
Moreover, this and other invariances lead to an additional bias in the objective of practical variational inference methods \cite{Kurle2022VBNN}, often resulting in very poor fits to the data \cite{Ghosh2018}. 

%% file: sec/method.tex
\section{Bayesian layerwise inference}\label{sec:methods}
We described NNs as a sequence of linear transformations of the features extracted from the preceding layers (cf.~\eqref{eq:neural_network_function}). 
Despite this simple algebraic structure, approximate inference is complicated by the significant multi-modality of the posterior distribution, as explained in Sec.~\ref{sec:BNN}. 
We introduce Bayesian layerwise inference (BALI) to address this problem: taking the layerwise linear model view one step further, we treat each layer as a distinct multivariate Bayesian linear regression model.
As a consequence, the local posterior of each layer is uni-modal by design. 
The main idea is that we can infer the layerwise posterior analytically if we have \emph{pseudo-targets} for the outputs of the linear transformation in each layer. 
Compared to the linear model described in Sec.~\ref{sec:background:MVBLR}, we additionally need to infer or estimate the unknown (latent) layerwise noise covariances. 
Fortunately, by using a conjugate prior over the noise covariance, inference in the multivariate Bayesian linear regression model with unknown noise covariance is also tractable and has a similarly simple analytical solution (Sec.~\ref{sec:method:inference}).
We then extend the inference methods to the mini-batch setting in Sec.~\ref{sec:method:adaptation}.
Finally, we propose gradient backpropagation as simple but effective method to obtain pseudo-targets in Sec.~\ref{sec:method:targets}.

\subsection{Layerwise model and inference}\label{sec:method:inference}
We consider simple fully-connected NNs as described by \eqref{eq:neural_network_function}. 
Since each layer computes just a linear transformation of its in-going features, we view the layer as a multivariate linear regression model, where $\bz_{n,l}$ are predictions for some target vectors $\by_{n,l}$. 
In contrast to Sec.~\ref{sec:background:MVBLR}, we now have unknown noise covariance matrices $\bSigma_l$, which we therefore model as additional latent random variables. 
We assume that the prior $p$ over the weights and covariance matrices factorises over the layers, and similarly, we approximate the posterior as a layerwise factorising distribution $q$:
\begin{equation}
	p\left( \{ \Weights_l \}_{l=1}^{L}, \{ \bSigma_l \}_{l=1}^{L} \right)
	= \prod_{l=1}^{L} p(\Weights_l, \bSigma_l),
	~~~~~
	q \left( \{ \Weights_l \}_{l=1}^{L}, \{ \bSigma_l \}_{l=1}^{L} \right)
	= \prod_{l=1}^{L} q(\Weights_l, \bSigma_l).
\end{equation}
In the rest of this section, we consider the layerwise models with local inference. 
Hence, we drop the layer index $l$ for better readability and refer to layer input and output dimensions as $D_\bx$ and $D_\by$. 
The conjugate prior for the layerwise multivariate linear regression model with unknown noise covariance is the matrix-normal inverse-Wishart distribution with density $p(\Weights, \bSigma) = p(\Weights | \bSigma)\, p(\bSigma)$,
\begin{equation}\label{eq:LMVBLR:prior}
	p(\Weights | \bSigma) = \mathcal{MN}(\bM_{0}, \bR_{0}, \bSigma), 
	~~~~~
	p(\bSigma) = \mathcal{W}^{-1}(\bU_0, \rmu_0),
\end{equation}
where $\mathcal{W}^{-1}$ denotes the inverse-Wishart distribution with scale matrix $\bU_0 \in \mathbb{R}^{D_\by \times D_\by}$ and a scalar degrees of freedom parameter $\rmu_0$. 
The density $p(\Weights | \bSigma)$ is identical to the prior in \eqref{eq:MVBLR:prior} in Sec.~\ref{sec:background:MVBLR}.
Due to the conjugacy of the above prior, the posterior $q(\Weights, \bSigma) \coloneqq p(\Weights, \bSigma | \bX, \bY)$ is analytically tractable and in the same family, i.e.\ 
$q(\Weights, \bSigma) = q(\Weights | \bSigma)\, q(\bSigma)$ with 
\begin{equation}
	q(\Weights | \bSigma) = \mathcal{MN}(\bM, \bR, \bSigma), 
	~~~~~
	q(\bSigma) = \mathcal{W}^{-1}(\bU, \rmu).
\end{equation}
The parameters of this posterior are given by (cf.~App.~\ref{app:mniw_posterior_natural_params})%
\begin{center}
\vspace{-2em}
\begin{subequations}\label{eq:method:layerwise_posterior_parameters}
\begin{minipage}[t]{0.38\linewidth}
\begin{align}
    \bR &= (\bR^{-1}_0 + \bX^\top \bX)^{-1}, \label{eq:method:layerwise_posterior_R}\\
    \bM &= \bR\, (\bR_0^{-1} \bM_0 + \bX^\top \bY), \label{eq:method:layerwise_posterior_mean}
\end{align}
\end{minipage}%
\hfill
\begin{minipage}[t]{0.58\linewidth}
\begin{align}
    \bU &= \bU_0 + \bM_0^\top \bR_0^{-1} \bM_0 - \bM^\top \bR^{-1} \bM + \bY^\top \bY, \label{eq:method:layerwise_posterior_U}\\
    \rmu &= \rmu_0 + N,
\end{align}
\end{minipage}
\end{subequations}
\vspace{-0.35em}
\end{center}
where $N$ is the dataset size, $\bX$ are the features and $\bY$ are the pseudo-targets of the layer (cf.~Sec.~\ref{sec:method:targets}). 

\subsection{Mini-batch inference through adaptation}\label{sec:method:adaptation}
Sec.~\ref{sec:method:inference} assumed that we compute the full-batch posterior, i.e.\ using features and pseudo-targets from the entire dataset. 
This is of course infeasible in practice for large datasets and models. 
For the mini-batch setting, we could consider the Bayesian online learning formulation
$
p(\weights | \Dataset_{1:t}) \propto p(\Dataset_{t} | \weights)\, p(\weights | \Dataset_{1:t-1}),
$
where $\Dataset_t \subset \Dataset$ is a random mini-batch at iteration $t$. 
However, this is problematic when iterating over multiple epochs, as we would revisit the data multiple times, and the posterior approximation would converge to a dirac distribution. 
Furthermore, the distribution over input features and pseudo-target depends on samples from the previous posterior approximation used for computing the pseudo-targets (cf.~\ref{sec:method:targets}). 
In other words, the sampling distribution of the pseudo-data exhibits a distribution shift during the learning process.

In order to account for this distribution shift and allowing to revisit the same data multiple times, we wish to gradually \emph{forget} contributions to the posterior that correspond to older features and pseudo-targets. 
To this end, we impose an exponential decay on the likelihood terms such that terms involving older features and pseudo-targets are down-weighted. 
In case of exponential family distributions, this can be achieved by using an exponential moving average (EMA) over terms corresponding to the natural parameters stemming from the likelihood (which involves the data). 
For our matrix-normal inverse-Wishart distribution, this corresponds to the terms $\bX^\top \bX$, $\bX^\top \bY$ and $\bY^\top \bY$ (cf.~App.~\ref{app:mniw_posterior_natural_params}), where $\bX$ and $\bY$ refer to the entire dataset of $N$ input features and pseudo-targets (cf.~\eqref{eq:method:layerwise_posterior_parameters}).
Now consider a random mini-batch of features $\bX_t \in \mathbb{R}^{B \times D_\bx}$ and pseudo-targets $\bY_t \in \mathbb{R}^{B \times D_\by}$ at iteration $t$. 
We estimate the natural parameters using the following EMA updates:
\begin{subequations}\label{eq:method:ema_statistics}
\begin{align}
\bPsi^{\mathrm{xx}}_t &= (1-\beta) \cdot \bPsi^{\mathrm{xx}}_{t-1} + \beta \cdot \frac{N}{B} \cdot \bX_t^\top \bX_t , \\
\bPsi^{\mathrm{xy}}_t &= (1-\beta) \cdot \bPsi^{\mathrm{xy}}_{t-1} + \beta  \cdot \frac{N}{B} \cdot \bX_t^\top \bY_t  , \\
\bPsi^{\mathrm{yy}}_t &= (1-\beta) \cdot \bPsi^{\mathrm{yy}}_{t-1} + \beta \cdot \frac{N}{B} \cdot \bY_t^\top \bY_t  
\end{align}
\end{subequations}
We multiply by the factor $\frac{N}{B}$, where $N$ is the dataset size and $B$ is the batch size, in order to estimate the likelihood terms (natural parameters) corresponding to the entire dataset. 
The hyperparameter $\beta \in [0, 1]$ is the update rate and $(1-\beta)$ is the rate at which old data is forgotten.
Since we initialise each of these terms with zero matrices, the estimate will be biased. 
However, we correct for the bias by dividing the estimates by the factor $b_t = 1 - (1-\beta)^{t}$, similarly to the bias correction in the Adam optimiser (cf.~\cite{Kingma2014Adam}).
Finally, we compute the layerwise posterior approximation at iteration $t$ by replacing the terms in \eqref{eq:method:layerwise_posterior_parameters} with the bias-corrected EMA estimates of the natural parameters, $\bX^\top \bX \approx \nicefrac{\bPsi^{\mathrm{xx}}_t}{b_t}$, $\bX^\top \bY \approx \nicefrac{\bPsi^{\mathrm{xy}}_t}{b_t}$ and $\bY^\top \bY \approx \nicefrac{\bPsi^\mathrm{yy}_t}{b_t}$.

\subsection{Target projection}\label{sec:method:targets}
Our posterior approximation with exact layerwise inference rests on the assumption that we have input features and pseudo-targets in every layer. 
While it is straightforward to obtain features from the \emph{forward} pass through the model, obtaining pseudo-targets must involve some form of information flow \emph{backward} from the actual targets. 
Perhaps the simplest but effective idea is to leverage gradient backpropagation, computing gradients of some objective function $\mathcal{L}\,$ w.r.t.\ the linear layer's outputs (not w.r.t.\ weights):
\begin{align}
\by_{n,t} 
&= \bz_{n,t} + \alpha \cdot \frac{\partial \mathcal{L}\left(\Dataset_t, \weights_{t}  \right)}{\partial \bz_{n,t}},
\end{align}
where we denote the set of weight samples from all layers by $\weights_{t} = \{ \Weights_{l,t} \}_{l=1}^{L}\,$. 
The weights used to compute the new pseudo-targets at iteration $t$ are drawn from the previous posterior approximation, i.e.\ $\Weights_{l,t} \sim q_{t-1}(\Weights_{l}|\bSigma_l)$, where $q_{t-1}$ is the (layerwise) posterior approximated with the natural parameter EMA from the previous iteration. 
Furthermore, $\Dataset_t$ denotes the random mini-batch used at iteration $t$, $\bz_{n,t}\,$ is the layer output from the forward pass (cf.~\eqref{eq:neural_network_function}) and the hyperparameter $\alpha$ is the step size. 
To simplify tuning the step size $\alpha$, we use gradient-updated targets that are invariant to re-scaling of the objective function. 
Inspired by the Adam optimiser and similarly to the previous section, we estimate the mean-squared gradients using an EMA:
\begin{align}
    \bPsi^{\mathrm{gg}}_t &= (1-\beta) \cdot \bPsi^{\mathrm{gg}}_{t-1} + \beta \cdot \frac{1}{B} \sum_{n=1}^{B} \mathbf{g}^2_{n,t}, 
\end{align}
where $\mathbf{g}_{n,t} = \frac{\partial \mathcal{L}\left(\Dataset_t, \weights_t  \right)}{\partial \bz_{n,t}}$ denotes the gradients for one of the $B$ samples in the mini-batch and the square is an element-wise multiplication. 
We then define the pseudo-targets as 
\begin{align}\label{eq:methods:pseudo_targets}
\by_{n,t} 
&= \bz_{n,t} + \alpha \cdot \frac{\mathbf{g}_{n,t}}{\sqrt{\nicefrac{\bPsi^{\mathrm{gg}}_t}{b_t}}} \,,
\end{align}
where the division is element-wise, i.e.\ per every output node, and the second division by the scalar $b_t$ is again to correct the bias in the EMA estimate (cf.~Sec.~\ref{sec:method:adaptation}).

We could sample the posterior ${q(\Weights, \bSigma) = q(\Weights | \bSigma)\, q(\bSigma)}$ (in each layer) by first sampling the noise covariance ${\bSigma \sim q(\bSigma)}$ and then sample weights ${\Weights \sim q(\Weights | \bSigma)}$. 
Here we instead use the most-likely noise covariance, i.e.\ the  mode of the inverse-Wishart distribution
\begin{align}\label{eq:inverse_wishart_mode}
    \bSigma = (\rmu + D_\by + 1)^{-1} \cdot \bU.
\end{align}
We then sample the weights $\Weights \sim q(\Weights | \bSigma)\,$ from the matrix-normal as follows: first draw $D_\bx \cdot D_\by$ samples from a standard normal distribution and stack the samples into a matrix $\mathbf{A} \in \mathbb{R}^{D_\bx \times D_\by}$. 
Then, transform these samples via the Cholesky factors of the two scale matrices as
\vspace{-0.2em}
\begin{align} \label{eq:weight_sample_matrix_normal}
    \Weights &= \bR^{\nicefrac{1}{2}} \mathbf{A} \bSigma^{\nicefrac{1}{2}}.
\end{align}
\vspace{-0.2em}
The resulting weight matrix is a sample from the matrix-normal posterior distribution.

\subsection{Algorithm and complexity}\label{sec:algorithm_and_complexity}
We draw the initial weights from a normal distribution, where the variance is a hyperparameter (close to $1.0$) and use \eqref{eq:inverse_wishart_mode} and \eqref{eq:weight_sample_matrix_normal} to sample the weights in all subsequent iterations.
The pseudo-algorithm for BALI is shown in \cref{alg:layerwise_mvblr} together with other closely related algorithms such as noisy K-FAC. 
The overall computational complexity of BALI is of order $O\big((D_{\bx}^2 + D_{\bx} D_{\by} + D^2_{\by}) B  + D_{\bx}^3 + D_{\by}^3 + D_{\bx}^2 D_{\by} + D_{\bx} D^2_{\by}  \big),$
where the terms involving batch size $B$ stem from the forward and backward pass as well as computing the EMA in \eqref{eq:method:ema_statistics}. The remaining terms are due to the matrix inversion of $\bR^{-1}$, Cholesky decomposition of $\bU$ and computing $\bM$ in \eqref{eq:method:layerwise_posterior_parameters}. 
See App.~\ref{app:computational_complexity} for an in-depth analysis and discussion.
We also derived an alternative formulation of the EMA update equations in App.~\ref{app:mean_updates}; while not used in this work, it provides several useful insights.

%% file: sec/related_work.tex
\section{Related work}\label{sec:related_work}
\textbf{Variational Inference.~}
One of the first steps towards developing BNNs was a variational method for computing derivatives while Gaussian noise is added to the weights \cite{Hinton1993a} (with the aim of improving generalisation).
In order to scale Bayesian inference to complex neural-network architectures, a gradient estimator for Gaussian posterior distributions was introduced in \cite{Graves2011}---with the tradeoff of this estimator being biased. 
This limitation was addressed in \cite{Blundell2015a} by applying the reparameterisation trick \cite{Kingma2013VAE}. 
The resulting algorithm is known as \textit{Bayes By Backprop} (BBB) and the basis for numerous approaches to learning BNNs \cite{Kingma2015LocalReparam, Ghosh2018, Farquhar2020Radial, Tomczak2021Collapsed} due to its simple implementation and robust optimisation performance.
Nevertheless, BBB often leads to poor empirical results \cite{Trippe2017, Ghosh2018}. 
This is further exacerbated by the fact that more restricted approximations often lead to better performance \cite{Dusenberry2020Rank1Factors, Swiatkowski2020, Tomczak2020LowRank}.
The crux of this problem lies in variational inference when applied to over-parameterised models, as it suffers from permutation and translation invariances present in deep NNs \cite{Kurle2022VBNN}.

\textbf{Natural-Gradient Variational Inference.~}
A solution to the above problem can be provided by natural gradient methods that use the GGN matrix in place of the Fisher information matrix. This can be interpreted as replacing the non-linear NN by a \emph{local linear} approximation, centered at the current weights \cite{Martens2020NG}. 
Popular algorithms are either based on Adam \cite{Kingma2014Adam} or on K-FAC \cite{Martens2015KFAC}.
Noisy Adam \cite{Zhang18NoisyKFAC}, Vadam \cite{khan2018fast}, and VOGN \cite{Osawa2019} use a diagonal approximation of the Fisher matrix to define the posterior covariance over the weights, which introduces a natural-gradient based update of the mean that is similar to Adam. 
Noisy K-FAC \cite{Zhang18NoisyKFAC} uses a Kronecker-factorised approximation of the Fisher matrix, as introduced by K-FAC. 
This allows decomposing the posterior covariance for each layer of a NN as a Kronecker product of two terms that define the scale matrices of a matrix-normal distribution. 
The resulting block-diagonal structure of the posterior imposes an independence w.r.t.\ different neural-network layers but captures more information than the diagonal covariance of the Adam-based approaches. 
In contrast to deterministic optimisers, these natural gradient variational inference methods additionally use adaptive weight noise in the update rule (see \cref{alg:noisy_adam,alg:vadam,alg:vogn,alg:noisy_kfac} in the appendix). 
Noisy K-KFAC shares many similarities with BALI:
In both algorithms, the update of $\bR$ is proportional to $\bX^\top \bX$ and only differs in the scaling of the prior precision and the EMA.
In contrast to Noisy K-FAC, our update of $\bM$ is \textit{not} proportional to $\bSigma$. 
This is due to the choice of a conjugate prior-distribution, which causes $\bSigma$ being canceled out (see \eqref{eq:MVBLR:mean}).
Another difference is that BALI assumes $\bSigma$ to be a random variable. 
Therefore, in contrast to Noisy K-FAC, we do not have a direct update rule for $\bSigma$ but for the parameters of $q(\bSigma)$---which we then use to define~$\bSigma$.

\textbf{Layerwise linear regression.~}
Interestingly, it has been shown that the effectiveness of (deterministic) K-FAC-based optimisation methods does not rely on the second Kronecker factor, i.e.\ the scale matrix $\bSigma$  \cite{Benzing2022FOOF}.
Instead, they interpret K-FAC as gradient-descent on neurons, similarly to how we explicitly defined the gradient-updated layer outputs as pseudo-targets. 
Their proposed (deterministic) optimisation method FOOF computes the weights as the solution to a (non-Bayesian) linear regression problem, similar to our mean updates. 
By contrast, we propose a probabilistic inference method, using the inferred covariance for sampling the weights and thus making probabilistic predictions. 

Bayesian layerwise inference has also been explored before from different perspective, using \emph{global inducing points} to define a variational posterior distribution \cite{Ober2021GlobalInducingPoint}. 
These inducing points are learned data points that are propagated through the entire NN, defining both pseudo-inputs and pseudo-targets for a Bayesian linear regression problem.  
The motivation and advantage of this approach is that it models correlations between layers. 
This is because this variational posterior has a layerwise conditional structure, where the distribution of the weights in some layer depends on the weights of all previous layers via the forward propagated inducing points. 
However, this approach is tractable only for a small number of inducing points, since all of these learned data points have to be propagated through the NN in order to sample from the posterior. 
By contrast, BALI does not require any learned inducing points and instead propagates the actual data through the network, defining the pseudo-targets of each layer as the outputs of the linear transformation updated by a gradient step. 
Posterior inference is then made tractable by using an EMA of mini-batch estimates of the posterior natural parameters, exploiting the property that the natural parameters of the posterior can be written as a sum over terms corresponding to individual data points or mini-batches. 

%% file: sec/experiments.tex
\section{Experiments}\label{sec:experiments}

\subsection{Synthetic data} \label{sec:experiments:toy}
\begin{figure}[htb!]
  \vspace{-1.0em}
  \centering
  \subfigure[Sines-trend dataset.]{
  	\includegraphics[width=.33\linewidth]{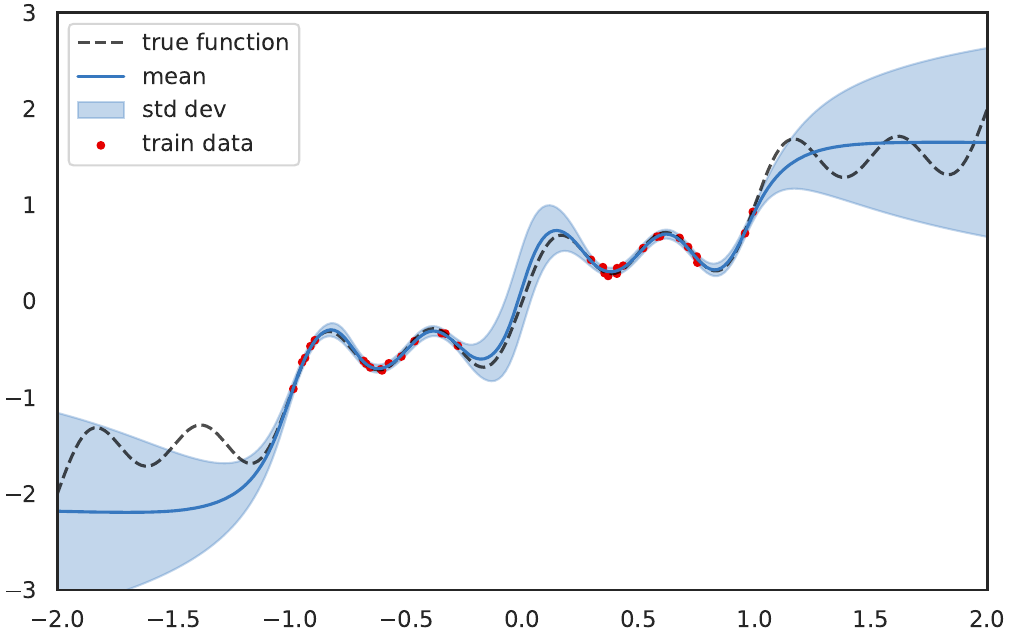}\label{fig:sine}
  	}
  \hfill
  \subfigure[Sinc dataset.]{
  \includegraphics[width=.33\linewidth]{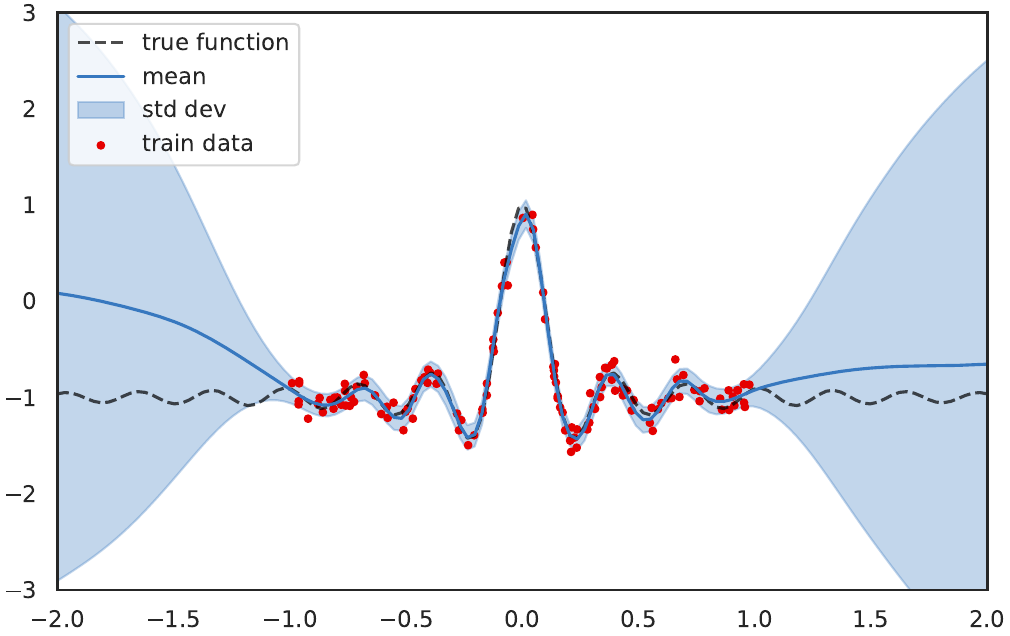}\label{fig:sinc}
  }
  \hfill
  \subfigure[Two-Moons dataset.]{
  \includegraphics[width=.275\linewidth]{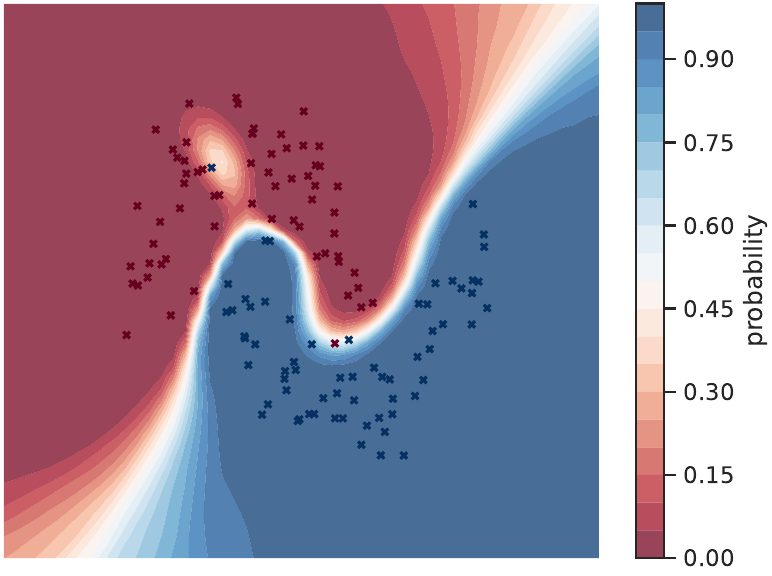}
  \label{fig:moon} 
  }
   \caption{Posterior-predictive of BALI on synthetic regression and classification datasets (cf.~Sec.~\ref{sec:experiments:toy}).}
  \label{fig:regression}
\end{figure}

We start by verifying that BALI is able to avoid the problem of underfitting in case of few data points compared the a larger model size and qualitatively assess the inter- and extra-polation behaviour of the posterior predictive distribution. 
To this end, we fit a NN on two regression and one classification dataset. 
For each of the 3 datasets, we used the same NN architecture, i.e.\ 3 hidden layers with 256 units each and tanh activation functions. 
The true function underlying the sines-trend data is the sum of two sine-functions and a linear component, and the sinc dataset was generated from a scaled and shifted sinc-function (see App.~\ref{app:experiment_details:synthetic} for details). 
For sines-trend, we generated 32 data points, sampling inputs uniformly in two disconnected regions and applied additive noise with standard deviation $0.02$ to the outputs. 
For sinc, we generated 128 data points using noise with standard deviation $0.1$. 
For classification, we consider the two-moons dataset, where we generated 128 data points with input standard deviation $0.2$. 
The predictive distributions are shown in Fig.~\ref{fig:regression}, showing that BALI leads to an excellent and smooth fit to the data with small uncertainty close to the data points and increased predictive uncertainty further away from the training data and decision boundary.

\subsection{Regression benchmark} 
\label{sec:experiments:regression}
We evaluate BALI on standard UCI regression datasets used in previous benchmarks of BNNs \cite{Hernandez2015PBP, Gal2016, Zhang18NoisyKFAC, khan2018fast}. 
Following the setup of these previous works, we evaluated the predictive performance of our method using $20$ random splits of the datasets, where $90 \%$ is used for training and $10 \%$ for testing. 
We also used the same model architecture as previous work, which is a NN with a single hidden layer of 50 units (100 for the protein structure dataset) and RELU activation function (see App.~\ref{app:experiment_details} for more details).
For regression with a Gaussian noise assumption, we simply use the inferred noise covariance matrix of the last layer to define the Gaussian likelihood and we use the log-likelihood as an objective function.
Furthermore, we use the true targets in the last layer rather than pseudo-targets. 
In Tab.~\ref{tab:uci}, we summarise the mean and standard errors for the test RMSE. The log-likelihood results are provided in Tab.~\ref{tab:results} of the supplementary material.
For the baselines NNG-MVG \cite{Zhang18NoisyKFAC}, VADAM \cite{khan2018fast}, BBB \cite{Blundell2015a} and PBP \cite{Hernandez2015PBP}, we use the results reported in the respective papers; for Monte-Carlo dropout (DO) \cite{Gal2016}, we used the author's implementation with improved results. 
It can be seen that BALI outperforms or has similar performance compared to the baselines. 
\begin{table}[tb!]
    \centering
    \caption{Averaged test RMSE for the regression benchmark (cf.~Sec.~\ref{sec:experiments:regression}). Lower is better.}
    \label{tab:uci}
    \begin{adjustbox}{width=\textwidth}
        \begin{tabular}{lccccccc}
            \toprule
            \textbf{Dataset} & \textbf{BALI} & \textbf{NNG-MVG} & \textbf{VADAM} & \textbf{BBB} & \textbf{PBP} & \textbf{DO} \\
            \midrule
            Yacht     & $\bold{\phantom{+}0.642 \pm 0.042}$ & $\phantom{+}0.979\pm0.077$ &  $\phantom{+}1.32 \pm 0.10$ & $\phantom{+}1.174\pm0.086$ & $\phantom{+}1.015 \pm 0.054$ & $\phantom{+} \bold{0.666 \pm 0.049}$ \\
            Concrete   & $\bold{\phantom{+}4.425 \pm 0.153}$ & $\phantom{+}5.019\pm0.127$ & $\phantom{+}6.85 \pm 0.09$ & $\phantom{+}5.678\pm0.087$ & $\phantom{+}5.667\pm0.093$ &  $\phantom{+}4.826 \pm 0.161$ \\
            Energy     & $\phantom{+}\bold{0.451 \pm 0.016}$ & $\phantom{+}0.485\pm0.023$ & $\phantom{+}1.55 \pm 0.08$ & $\phantom{+}0.565\pm0.018$ & $\phantom{+}1.804\pm0.048$ &  $\phantom{+}0.539 \pm 0.014$\\
            Redwine    & $\phantom{+}0.676 \pm 0.016$ & $\phantom{+}0.637\pm0.011$ & $\phantom{+}0.66 \pm 0.01$ & $\phantom{+}0.643\pm0.012$ & $\phantom{+}0.635\pm0.008 $ & $\bold{\phantom{+}0.622 \pm 0.008}$ \\
            Kin8nm     & $\phantom{+}\bold{0.075 \pm 0.001}$ & $\phantom{+} \bold{0.076\pm0.001}$ & $\phantom{+}0.10 \pm 0.00$ & $\phantom{+}0.080\pm0.001$ & $\phantom{+}0.098\pm0.001$ & $\phantom{+}0.079 \pm 0.001$\\
            Powerplant & $\phantom{+}\bold{3.882 \pm 0.026}$ & $\phantom{+}\bold{3.886\pm0.041}$ & $\phantom{+}4.28 \pm 0.03$ & $\phantom{+}4.023\pm0.036$ & $\phantom{+}4.124\pm0.035$ & $\phantom{+}4.014 \pm 0.037$\\
            Protein & $\phantom{+} 4.129 \pm 0.016$ & $\phantom{+} \bold{4.097 \pm 0.009} $ & $\phantom{+} - $ & $\phantom{+} 4.321 \pm 0.017$ & $\phantom{+} 4.732 \pm 0.013$ & $\phantom{+}4.274 \pm 0.021$\\
            \bottomrule
        \end{tabular}
    \end{adjustbox}
\end{table}

\subsection{Classification benchmark}\label{sec:experiments:classification}
Next, we evaluate BALI on the classification datasets Spambase, Diabetes Health Indicators and Magic Telescope from the UCI repository as well as MNIST and FashionMNIST. 
We compare BALI to Monte-Carlo dropout (DO) using dropout probability $p=0.1$ and a deterministic model learned via maximum a posteriori (MAP) estimation. 
For both DO and MAP, the weights are learned using the ADAM optimiser. 
The NN consists of $2$ hidden layers with $256$ units and a Leaky-Tanh activation function, $f(x) = \mathrm{tanh}(x) + 0.1 x$. 
For details regarding the hyperparameters, see App.~\ref{app:experiment_details}. 
We evaluate the predictive performance in terms of classification accuracy (ACC) and expected calibration error (ECE). 
In case of binary classification, we also compute the area under the ROC curve (AUC). 
The results are summarised in Tab.~\ref{tab:classification_benchmark}, showing that BALI obtains similar accuracy as Monte Carlo DO and is better calibrated in 4 out of 5 cases. 
\begin{table}[tb!]
    \centering
    \caption{Comparison of classification and calibration metrics for classification benchmark (Sec.~\ref{sec:experiments:classification}).}
    \label{tab:classification_benchmark}
    \begin{adjustbox}{width=1.0\textwidth}
    \begin{tabular}{lcccccc}
        \toprule
        & \multicolumn{2}{c}{\textbf{ACC} $\uparrow$ } & \multicolumn{2}{c}{\textbf{ECE} $\downarrow$} & \multicolumn{2}{c}{\textbf{AUC} $\uparrow$} \\
        \cmidrule(lr){2-3} \cmidrule(lr){4-5} \cmidrule(lr){6-7}
        \textbf{Dataset} & \textbf{BALI} & \textbf{DO} & \textbf{BALI} &  \textbf{DO} & \textbf{BALI} & \textbf{DO} \\
        \midrule
        Diabetes   & $\bold{0.867 \pm 0.000}$ & $\bold{0.867 \pm 0.000}$ & $\bold{0.004 \pm 0.000}$ & $0.006 \pm 0.000$ & $\bold{0.831 \pm 0.000}$ & $\bold{0.831 \pm 0.000}$ \\
        Telescope   & $\bold{0.883 \pm 0.001}$ & $0.874 \pm 0.001$ & $\bold{0.015 \pm 0.001}$ & $0.040 \pm 0.003$ & $\bold{0.938 \pm 0.001}$ & $0.934 \pm 0.002$ \\
        Spam  & $0.939 \pm 0.005$ & $\bold{0.945 \pm 0.003}$ & $0.029 \pm 0.003$ & $\bold{0.022 \pm 0.002}$ & $0.980 \pm 0.002$ & $\bold{0.985 \pm 0.001}$ \\
        MNIST    & $\bold{0.986 \pm 0.000}$ & $\bold{0.986 \pm 0.000}$ & $\bold{0.003 \pm 0.000}$ & $0.006 \pm 0.000$ & - & - \\
        F-MNIST  & $\bold{0.905 \pm 0.001}$ & $0.903 \pm 0.000$ & $\bold{0.008 \pm 0.001}$ & $\bold{0.008 \pm 0.000}$ & - & - \\
        \bottomrule
    \end{tabular}
    \end{adjustbox}
\end{table}

\subsection{Convergence speed}
\begin{figure}[htb!]
  \centering
  \subfigure[FashionMNIST.]{\includegraphics[width=0.245\textwidth]{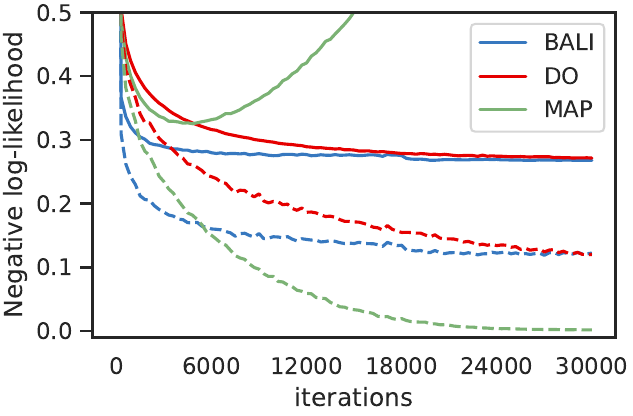}}
  \subfigure[MNIST.]{\includegraphics[width=0.245\textwidth]{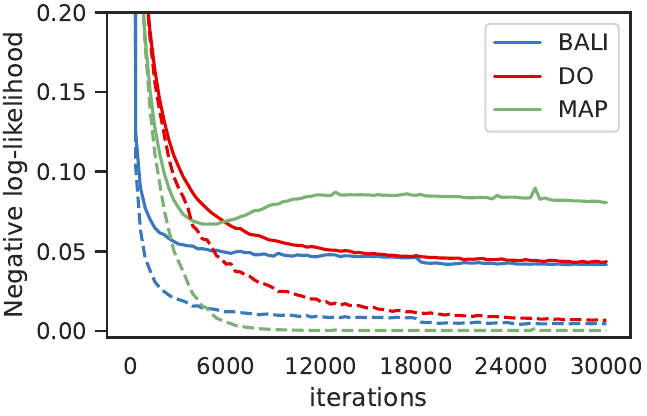}}
  \subfigure[Telescope.]{\includegraphics[width=0.245\textwidth]{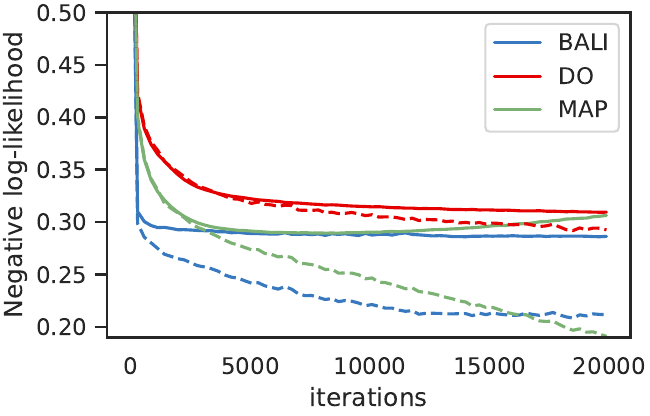}}
  \subfigure[Spam.]{\includegraphics[width=0.245\textwidth]{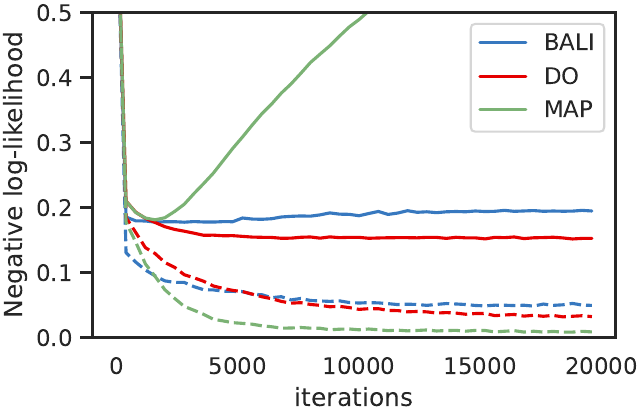}}
  \caption{
  NLL averaged over test set (solid line) and training set (dashed line), plotted over iterations. 
  }
  \label{fig:learning_curves}
\end{figure}

Since BALI performs locally exact posterior updates using backpropagated pseudo-targets, we expect the inference algorithm to converge in significantly fewer iterations compared to standard gradient-based methods that optimise the weights directly. 
We validate this hypothesis by comparing the negative log-predictive likelihood (NLL) of BALI to Monte Carlo DO (with dropout probability $0.1$) and deterministic MAP estimation. 
On MNIST and FashionMNIST, we train each method for $1000$ epochs using batch size $2000$, resulting in $30000$ iterations, and average over $3$ random seeds.
On Telescope and Magic, we train for $20000$ iterations with batch size $2048$ and average over $5$ seeds. 
The NLL is evaluated every $10$ epochs, using $128$ random samples from the current posterior and dropout masks, respectively.
These per-data log-likelihood values are averaged over the entire test set and $10000$ random data points from the training set, respectively. 
As can be seen in Fig.~\ref{fig:learning_curves}, BALI converges in significantly fewer iterations compared to DO and MAP.
It must be noted that the fast convergence of BALI also leads to a gap between training and test performance more quickly. 

\vspace{-1em}
\subsection{Out-of-distribution detection}\label{sec:exp:classification:oodd}
\begin{wrapfigure}{r}{0.51\textwidth} 
	\vspace{-4em}
	\centering
	\subfigure[FashionMNIST (ID).]{\includegraphics[width=0.25\textwidth]{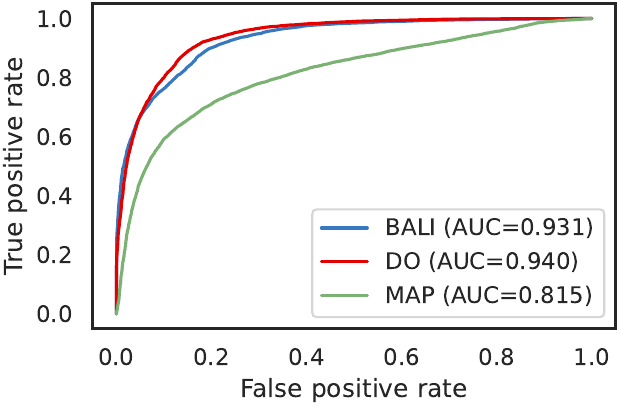}}
	\subfigure[MNIST (ID).]{\includegraphics[width=0.25\textwidth]{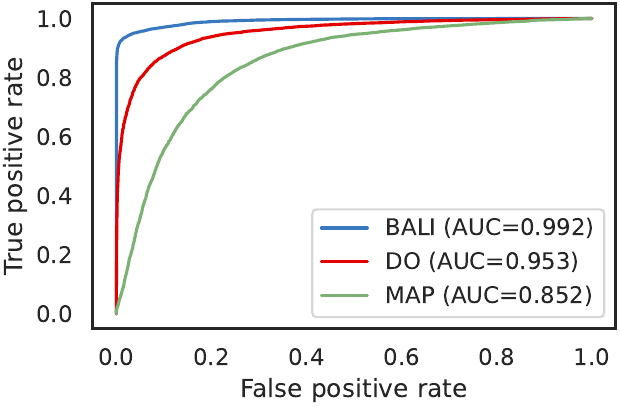}}
	\hfill
	\label{fig:ood:fashion_vs_mnist_ROC_AUC}
	\vspace{-.70em}
	\caption{OOD detection ROC curves using the negative entropy of the predictive distribution to distinguish ID from OOD. The model is trained on the ID dataset and the respective other set is used as OOD data. True positives are ID data classified as ID, false positives are ID data classified as OOD. 
	}
	\vspace{0em}
\end{wrapfigure}
We compare the out-of-distribution (OOD) detection performance of BALI against Monte Carlo DO and standard MAP estimation on the MNIST and FashionMNIST datasets, where we train on the in-distribution (ID) dataset and treat the respective other set as OOD. 
We then classify ID and OOD data based on a threshold of the negative entropy of the predictive distribution. 
We visualise the receiver operating characteristic (ROC) curves resulting from these thresholds in Fig.~\ref{fig:ood:fashion_vs_mnist_ROC_AUC}. 
It can be seen that BALI performs very similar to DO when FashionMNIST is used as ID data and performs significantly better (with almost perfect detection) for MNIST (ID) vs. FashionMNIST (OOD). 

%% file: sec/conclusion.tex
\section{Summary and Outlook}
We introduced BALI, a new probabilistic method for learning NNs by treating layers as local linear regression models. 
The inputs and targets for the layerwise models are defined as the features extracted from the preceding layers and gradient-updated outputs, respectively. 
The resulting layerwise posterior is a matrix-normal distribution with a Kronecker-factorised covariance matrix that can be inverted efficiently. 
We then extended our approach to the mini-batch setting by gradually forgetting likelihood terms corresponding to old mini-batches. 
Experiments on synthetic data confirmed that, unlike standard variational inference methods, BALI avoids underfitting and leads to increased uncertainty in input regions far from the training data. 
On common regression, classification and out-of-distribution benchmark datasets, BALI performed similar or better than state-of-the-art baselines.

By using gradient updates to compute pseudo-targets, we arrived at an algorithm that is similar to noisy natural gradient descent using the K-FAC approximation \cite{Zhang18NoisyKFAC}. 
Besides several algorithmic differences described in Sec.~\ref{sec:related_work}, the main conceptional difference stems from viewing learning as local Bayesian inference using layerwise pseudo-targets. 
This crucial difference opens the door for variants of BALI with alternative methods for computing the pseudo-targets: for instance, approximating and sampling from the posterior over hidden nodes similarly to Gibbs sampling, using Monte Carlo methods, or predicting target representations with other neural networks.

On the other hand, BALI has limitations worth addressing in future work. 
For instance, the scope of this work is limited to small models and only fully-connected layers. 
However, modern NN architectures are composed of convolutional, recurrent, and attention layers with vastly more parameters, which are trained in a distributed manner (see App.~\ref{app:computational_complexity} for a discussion). 
Furthermore, BALI has no momentum term like most gradient-based methods, as that does not naturally follow from the layerwise inference perspective.
Finally, we found BALI to be sensitive to hyperparameter choices, similar to KFAC-based methods.
A promising research direction therefore concerns defining robust heuristics to find the hyperparameters. 

%% file: sec/appendix/kronecker_properties.tex
 \section{Properties of the Kronecker product}\label{app:kron}
The Kronecker product of the matrix $\bA \in \mathbb{R}^{k\times l}$ with the matrix $\bB \in \mathbb{R}^{m \times n}$ is defined as
\begin{equation}\label{eq:kronecker:definition}
    \bA \kron \bB = 
    \begin{bmatrix}
    \bA_{11} \bB & \dots  & \bA_{1 l} \bB \\
    \vdots  & \ddots  & \vdots \\
    \bA_{k 1} \bB & \dots & \bA_{k l} \bB
    \end{bmatrix}
\end{equation}

Here we list several basic and well-known properties of the Kronecker product that are used in this paper. 
Proofs can be found e.g.\ in \cite{Loan2000Kron} and \cite{Zhang2013Kronecker}. 
\begin{property}[Distributivity]\label{thm:kronecker:distributive}
$\forall \bA, \bB \in \mathbb{R}^{k \times l}, \bC \in \mathbb{R}^{o \times p}:$
\begin{equation}
    (\bA + \bB) \kron \bC = \bA \kron \bC + \bB \kron \bC.
\end{equation}
\end{property}

\begin{property}[Associativity]\label{thm:kronecker:associativity}
$\forall \bA \in \mathbb{R}^{k \times l}, \bB \in \mathbb{R}^{m \times n}, \bC \in \mathbb{R}^{o \times p}:$
\begin{equation}
    (\bA \kron \bB) \kron \bC = \bA \kron (\bB \kron \bC).
\end{equation}
\end{property}

\begin{property}[Inverse]\label{thm:kronecker:inverse}
If $\bA \in \mathbb{R}^{k \times k}$ and $\bB \in \mathbb{R}^{m \times m}$ are non-singular, then
\begin{equation}
    (\bA \kron \bB)^{-1} = \bA^{-1} \kron \bB^{-1}.
\end{equation}
\end{property}

\begin{property}[Transposition]\label{thm:kronecker:transposition}
$\forall \bA \in \mathbb{R}^{k \times l}, \bB \in \mathbb{R}^{m \times n}:$
\begin{equation}
    (\bA \kron \bB)^{\top} = \bA^{\top} \kron \bB^{\top}.
\end{equation}
\end{property}

\begin{property}[Mixed product]\label{thm:kronecker:mixed}
$\forall \bA \in \mathbb{R}^{k \times l}, \bB \in \mathbb{R}^{m \times n}, \bC \in \mathbb{R}^{l \times p}, \bD \in \mathbb{R}^{n, r}:$
\begin{equation}
    (\bA \kron \bB)(\bC \kron \bD) = (\bA \bC) \kron (\bB \bD).
\end{equation}
\end{property}

As a special case of the mixed product property, it follows that, for $\bB=1$ and $\bD \in \mathbb{R}^{1 \times r}$,
\begin{equation}
    \bA (\bC \kron \mathbf{d}^{\top}) = (\bA \bC) \kron \mathbf{d}^{\top}.
\end{equation}
If we further have, $\bC = \mathbf{I}_{l}$, then
\begin{equation}
    \bA (\mathbf{I}_{l} \kron \mathbf{d}^{\top}) = \bA \kron \mathbf{d}^{\top}.
\end{equation}

\begin{property}[Vectorisation of matrix multiplication]\label{thm:kronecker:Vectorisation}
\begin{align}\label{eq:kron_vectorisation_identity}
    \mathrm{vec} \left( \bA \bX \bB \right)
    &= (\bB^{\top} \kron \bA) \mathrm{vec} \left( \bX \right) 
\end{align}
\end{property}

%% file: sec/appendix/posterior_parameters.tex
\section{Multivariate Bayesian linear regression (known $\bSigma$): posterior parameters}\label{sec:appendix:MVBLR}
Here, we show that the natural parameters of the multivariate Bayesian linear regression model are given by \eqref{eq:MVBLR:natural_params} and the distribution parameters are given by \eqref{eq:MVBLR:distribution_params}.

\paragraph{Posterior precision, \eqref{eq:MVBLR:precision}.} 
We want to show that 
\begin{align*}
    \bV^{-1} 
    = \bV_0^{-1} + \bV_{1:N}^{-1} 
    &= \bV_0^{-1} + \sum_{n=1}^{N} \bar{\bX}_n^{\top} \bSigma^{-1} \bar{\bX}_n \\
    &= \bSigma^{-1} \kron (\bR_0^{-1} + \bX^{\top} \bX).
\end{align*}
First, consider the prior precision term, which is obtained easily via the inverse property \ref{thm:kronecker:inverse},
\begin{align}\label{eq:kron_prior_precision}
    \bV_0^{-1} = \bSigma^{-1} \kron \bR_0^{-1}.
\end{align}
Next, consider the $N$ precision terms, $\bar{\bX}_n^{\top} \bSigma^{-1} \bar{\bX}_n = (\mathbf{I}_{D_{\by}} \otimes \bx_n^{\top})^{\top} \bSigma^{-1} (\mathbf{I}_{D_{\by}} \otimes \bx_n^{\top})$, stemming from the likelihood. 
Using the mixed product property \ref{thm:kronecker:mixed} with $\bA=\bSigma^{-1}, \bB=\mathbf{I}_1 = 1, \bC=\mathbf{I}_{D_{\by}}, \bD=\bx_n^{\top}$:
\begin{align}
\bSigma^{-1} (\mathbf{I}_{D_{\by}} \otimes \bx_n^{\top}) 
&= (\bSigma^{-1} \kron \mathbf{I}_1) (\mathbf{I}_{D_{\by}} \otimes \bx_n^{\top}) \\
&= (\bSigma^{-1} \mathbf{I}_{D_{\by}}) \kron (\mathbf{I}_1 \bx_n^{\top}) 
= \bSigma^{-1} \kron \bx_n^{\top}.
\end{align}
Using the transposition property \ref{thm:kronecker:transposition} and the mixed product property, we have
\begin{align}
    (\mathbf{I}_{D_{\by}} \otimes \bx_n^{\top})^{\top} (\bSigma^{-1} \kron \bx_n^{\top})
    &= (\mathbf{I}_{D_{\by}} \otimes \bx_n) (\bSigma^{-1} \kron \bx_n^{\top}) \\
    &= (\mathbf{I}_{D_{\by}} \bSigma^{-1}) \kron (\bx_n \bx_n^{\top}) 
    = \bSigma^{-1}  \kron \bx_n \bx_n^{\top}.
\end{align}
Due to linearity of the sum and since $\sum_{n=1}^{N} \bx_n \bx_n^{\top} = \bX^{\top} \bX$,
\begin{align}
    \bV_{1:N}^{-1} 
    = \sum_{n=1}^{N} \bSigma^{-1}  \kron \bx_n \bx_n^{\top}  
    &= \bSigma^{-1} \kron \sum_{n=1}^{N} \left( \bx_n \bx_n^{\top} \right) \\
    &= \bSigma^{-1}  \kron \left(\bX^{\top} \bX \right).
\end{align}
Finally, it is easy to see that
\begin{align}
    \bV^{-1}  = \bV_0^{-1} + \bV_{1:N}^{-1} 
    &= \left( \bSigma^{-1} \kron \bR_0^{-1} \right) +  \left( \bSigma^{-1}  + \bX^{\top} \bX \right) \\
    &= \bSigma^{-1}  \kron \left( \bR_0^{-1} + \bX^{\top} \bX \right)
\end{align}
since both precision matrices use the same factor $\bSigma^{-1}$ in their respective Kronecker product.

\paragraph{Posterior precision-mean, \eqref{eq:MVBLR:precision-mean}.}
We want to show that
\begin{align*}
    \Eta &= \Eta_0 + \Eta_{1:N} = \bR_0^{-1} \bM_0 \bSigma^{-1} + \bX^{\top} \bY \bSigma^{-1}.
\end{align*}
We start with the prior precision-mean in vectorised form $\boldsymbol{\eta}_0$ and as a matrix $\Eta_0$ . 
Using property \ref{thm:kronecker:Vectorisation},
\begin{align}
    \boldsymbol{\eta}_0 
    = \bV_0^{-1} \bm_0 
    &= (\bSigma^{-1} \kron \bR_0^{-1}) \textrm{vec}(\bM_0) \\
    &= \textrm{vec}(\bR_0^{-1} \bM_0 \bSigma^{-1}).
\end{align}
Hence, we can see that 
\begin{align}
    \Eta_0 = \bR_0^{-1} \bM_0 \bSigma^{-1}.
\end{align}

Next, consider the $N$ precision-mean terms $\bar{\bX}^T_n \bSigma^{-1} \by_n = (\mathbf{I}_{D_{\by}} \otimes \bx_n^{\top})^{\top} \bSigma^{-1} \by_n$, stemming from the likelihood.
Using the transposition property \ref{thm:kronecker:transposition}, the property $\mathrm{vec}(\ba) = \mathrm{vec}(\ba^{\top})$, and property \ref{thm:kronecker:Vectorisation}, we have
\begin{align}
(\mathbf{I}_{D_{\by}} \kron \bx_n) \mathrm{vec}\left( \bSigma^{-1} \by_n \right) 
&=(\mathbf{I}_{D_{\by}} \kron \bx_n) \mathrm{vec}\left( \by_n^{\top} \bSigma^{-1} \right) \\
&= \mathrm{vec} \left( \bx_n \by_n^{\top} \bSigma^{-1} \mathbf{I}_{D_{\by}} \right) 
= \mathrm{vec} \left( \bx_n \by_n^{\top} \bSigma^{-1} \right).
\end{align}
Again, we can identify the precision-mean in matrix shape as $\Eta_{n} = \bx_n \by_n^{\top} \bSigma^{-1}$.
Since the vectorisation and the sum are linear and the order is thus interchangeable, we have
\begin{align}
    \boldsymbol{\eta}_{1:N}
    &=\sum_{n=1}^{N} \mathrm{vec} \left( \bx_n \by_n^{\top} \bSigma^{-1} \right) \\
    &= \mathrm{vec} \left(\sum_{n=1}^{N} \bx_n \by_n^{\top}  \bSigma^{-1} \right) 
    = \mathrm{vec} \left( \bX^{\top} \bY \bSigma^{-1} \right).
\end{align}
Putting the results for $\Eta_0$ and $\Eta_{1:N}$ together and noting that 
\begin{align}
    \boldsymbol{\eta} &= \mathrm{vec}(\Eta_0) + \mathrm{vec}(\Eta_{1:N}) = \mathrm{vec}(\Eta_0 + \Eta_{1:N}),
\end{align}
we can thus identify the precision-means in matrix form as
\begin{align}
    \Eta = \Eta_0 + \Eta_{1:N} 
    &= \bR_0^{-1} \bM_0 \bSigma^{-1} + \bX^{\top} \bY \bSigma^{-1}.
\end{align}

\paragraph{Posterior covariance.}
Using the inverse property \ref{thm:kronecker:inverse}, the posterior covariance matrix is given by
\begin{align}
\bV & = \bSigma \kron \bR = \bSigma \kron \left(\bR_0^{-1} + \bX^{\top} \bX \right)^{-1},
\end{align} 
with $\bR^{-1} = \bR_0^{-1} + \bR_{1:N}^{-1}$ and $\bR_{1:N}^{-1} = \bX^{\top} \bX $ as in the main text. 
Note that inverting the posterior precision matrix $\bV^{-1} \in \mathbb{R}^{D_{\bx} \cdot D_{\bx} \times D_{\bx} \cdot D_{\bx}}$ only has computation complexity of the order $O(D_{\bx}^{3} + D_{\bx}^{3})$ since we only need to invert the factors of the Kronecker product, which is comparable to the matrix-matrix multiplications of the forward pass.

\paragraph{Posterior mean, \eqref{eq:MVBLR:mean}.}
Finally, we compute the posterior means in vectorised and matrix form. 
Using property \ref{thm:kronecker:Vectorisation}, we have
\begin{align}
\bm = \bV \boldsymbol{\eta} 
= \left(\bSigma \kron \bR \right) \mathrm{vec}\left(\Eta \right) 
= \mathrm{vec}\left(\bR \Eta \bSigma \right).
\end{align}
Thus, the means in matrix shape are
\begin{align}
\bM = \bR \Eta \bSigma 
&= \left(\bR_0^{-1} + \bX^{\top} \bX \right)^{-1} \left( \bR_0^{-1} \bM_0 \bSigma^{-1} + \bX^{\top} \bY \bSigma^{-1} \right) \bSigma \\
&= \left(\bR_0^{-1} + \bX^{\top} \bX \right)^{-1} (\bR_0^{-1} \bM_0 + \bX^{\top} \bY ).
\end{align}

%% file: sec/appendix/natural_parameters.tex
\section{Matrix-normal-inverse-Wishart distribution posterior}\label{app:mniw_posterior_natural_params}
We use a prior that is conjugate to the likelihood function of the layerwise linear regression model. 
The conjugate prior is an exponential family distribution and has the same structure as the likelihood function. 
Taking the product between likelihood and prior corresponds to a sum of their respective natural parameters, thus resulting in the natural parameters of the posterior distribution. 
In this section, we write the prior and likelihood as an exponential-family distribution in canonical form, i.e.\ as
\begin{align}
    f(\bx|\btheta) = h(\bx) g(\btheta) \exp\left( \btheta^{\top} \boldsymbol{T}(\bx) \right)
\end{align}
where $\btheta$ are the natural parameters, $\boldsymbol{T}(\bx)$ are the sufficient statistics of the data $\bx$, and $\btheta^{\top} \boldsymbol{T}(\bx) = \sum_{k=1}^K \theta_k \cdot T_k(\bx)$ is the inner product of the vector-valued parameters and statistics.

\paragraph{Inverse-Wishart prior covariance matrix.} 
The prior distribution has a conditional structure $p(\bSigma, \Weights) = p(\bSigma) p(\Weights | \bSigma)$. 
The prior over covariance matrices follows an inverse-Wishart distribution
\begin{align}
    \bSigma \sim \mathcal{W}^{-1}(\bU_0, u_0),
\end{align}
with hyperparameters $\bU_0$ and $u_0$. 
We transform the density function into the canonical form, 
\begin{align}
\begin{split}
    p(\bSigma; \bU_0, u_0) 
    &= \frac{|\bU_0|^{\nicefrac{u_0}{2}}}{2^{\nicefrac{D_{\by} \cdot u_0}{2}} \cdot \Gamma_{D_{\by}}\left(\frac{u_0}{2}\right)} 
    |\bSigma|^{-\frac{u_0 + D_{\by} + 1}{2}} \exp{ \left( -\frac{1}{2} \text{tr}\left[\bU_0 \bSigma^{-1} \right] \right) } \\
    &\propto  \exp{\left(
    -\frac{u_0 + D_{\by} + 1}{2} \cdot \log |\bSigma| 
    +\text{tr}\left[ -\frac{1}{2} \bU_0 \bSigma^{-1} \right] 
    \right)} \\
    &=  \exp{\left(
    -\frac{u_0 + D_{\by} + 1}{2} \cdot \log |\bSigma| 
    + \text{vec}\Big(-\frac{1}{2} \bU_0 \Big)^{\top} \text{vec}\Big(\bSigma^{-1}\Big) 
    \right)},
\end{split}
\end{align}
where $D_{\by}$ is the dimension of the covariance matrix, $|\cdot|$ denotes the determinant and $\Gamma_{D_{\by}}$ is the multivariate gamma function. 
From the last line, we can directly read off the natural parameters as $-\frac{u_0 + D_{\by} + 1}{2}$ and $- \frac{1}{2} \bU_0$ with the corresponding sufficient statistics $\log |\bSigma|$ and $\bSigma^{-1}$. 
We write the second natural parameter in matrix form, noting that $\bU_0 = U_0^{\top}$ and the trace of the matrix multiplication can be written as the inner product between the vectorised matrices, i.e.\
$$\text{tr}[\bU_0 \bSigma^{-1}] = \text{tr}[\bU_0^{\top} \bSigma^{-1}] = \text{vec}(\bU_0)^{\top} \text{vec}(\bSigma^{-1}).$$

\paragraph{Matrix-normal prior weights matrix.}
The prior over the neural network weights follows a matrix-normal distribution 
\begin{align}
    \Weights \sim \mathcal{MN}(\bM_0, \bR_0, \bSigma).
\end{align}
This density is conditional on the matrix-variate variable $\bSigma$ and has hyperparameters $\bM_0$ and $\bR_0$. 
As before, we bring this (conditional) density into the canonical form,
\small
\begin{align}
\begin{split}
    &p(\Weights | \bSigma; \bM_0, \bR_0)
    = \frac{1}{(2\pi)^{\nicefrac{D_{\by} \cdot D_{\bx}}{2}} |\bSigma|^{\nicefrac{D_{\bx}}{2}} |\bR_0|^{\nicefrac{D_{\by}}{2}}} \exp\left(-\frac{1}{2} \text{tr}\left[(\Weights - \bM_0)^{\top} \bR_0^{-1} (\Weights - \bM_0) \bSigma^{-1} \right]\right) \\
    &\propto \exp\left(-\frac{D_{\bx}}{2} \cdot \log |\bSigma| -\frac{1}{2} \text{tr}\left[ (\Weights - \bM_0)^{\top} \bR_0^{-1} (\Weights - \bM_0) \bSigma^{-1} \right]\right) \\
    &= \exp\left(-\frac{D_{\bx}}{2} \cdot \log |\bSigma| - \frac{1}{2} \text{tr}\left[ \left(\Weights^{\top} \bR_0^{-1} \Weights + \bM_0^{\top} \bR_0^{-1} \bM_0 - \Weights^{\top} \bR_0^{-1} \bM_0 - \bM_0^{\top} \bR_0^{-1} \Weights  \right)\bSigma^{-1} \right]\right) \\
    &= \exp\left( 
    -\frac{D_{\bx}}{2} \cdot \log |\bSigma| +
    \text{tr}\left[ -\frac{1}{2}\bR_0^{-1} \Weights \bSigma^{-1} \Weights^{\top} \right]
    + \text{tr}\left[-\frac{1}{2} \bM_0^{\top} \bR_0^{-1} \bM_0 \bSigma^{-1} \right]
    + \text{tr}\Big[\bR_0^{-1} \bM_0 \bSigma^{-1} \Weights^{\top}  \Big]\right),
\end{split}
\end{align}
\normalsize
where $D_{\by}$ is again the number of outputs and $D_{\bx}$ is the number of inputs. 
Hence, the natural parameters are $-\frac{D_{\bx}}{2}$, $-\frac{1}{2} \bR_0^{-1}$, $-\frac{1}{2} \bM_0^{\top} \bR_0^{-1} \bM_0$, and $\bR_0^{-1} \bM_0$, with the corresponding sufficient statistics $\log |\bSigma|$, $\Weights \bSigma^{-1} \Weights^{\top}$, $\bSigma^{-1}$, and $\bSigma^{-1} \Weights^{\top}$.

\paragraph{Multivariate-normal likelihood.}
The layerwise per-data likelihood is a multivariate normal distribution
\begin{align}
    \by_n \sim \mathcal{N}(\Weights^{\top} \bx_n, \bSigma).
\end{align}
This density is conditional on the matrix-variate variables $\bSigma$ and $\Weights$. 
Next, we express the per-data likelihood as a function of the covariance matrix and weights, thereby exposing the natural parameters stemming from the likelihood:
\small
\begin{align}
\begin{split}
    &p(\by_n | \bx_n, \bSigma, \Weights) 
    = \frac{1}{(2\pi)^{\nicefrac{D_{\by}}{2}} |\bSigma|^{\nicefrac{1}{2}}} \exp{\left(
    -\frac{1}{2} \left(\Weights^{\top} \bx_n - \by_n \right)^{\top} \bSigma^{-1} \left(\Weights^{\top} \bx_n - \by_n \right) \right)} \\
    &\propto \exp{\left( - \frac{1}{2} \log |\bSigma|
    -\frac{1}{2} \text{tr}\left[ \left(\Weights^{\top} \bx_n - \by_n \right)^{\top} \bSigma^{-1} \left(\Weights^{\top} \bx_n - \by_n \right) \right] \right)} \\
    &= \exp{\left( 
    - \frac{1}{2} \log |\bSigma|
    -\frac{1}{2} \text{tr}\left[ \left(\Weights^{\top} \bx_n - \by_n \right) \left(\Weights^{\top} \bx_n - \by_n \right)^{\top} \bSigma^{-1}  \right] \right)} \\
    &= \exp{\left( 
    - \frac{1}{2} \log |\bSigma|
    -\frac{1}{2} \text{tr}\left[ \left(\Weights^{\top} \bx_n \bx_n^{\top} \Weights  + \by_n \by_n^{\top} - \Weights^{\top} \bx_n \by_n^{\top} - \by_n \bx_n^{\top} \Weights \right) \bSigma^{-1}  \right] \right)} \\
    &= \exp{\left( 
    - \frac{1}{2} \log |\bSigma|
    + \text{tr}\left[ -\frac{1}{2} \Weights^{\top} \bx_n \bx_n^{\top} \Weights  \bSigma^{-1} \right]  + \text{tr}\left[ -\frac{1}{2}  \by_n \by_n^{\top}  \bSigma^{-1} \right] + \text{tr}\left[ \Weights^{\top} \bx_n \by_n^{\top}  \bSigma^{-1} \right] \right)} \\
    &= \exp{\left( 
    - \frac{1}{2} \log |\bSigma|
    + \text{tr}\left[ -\frac{1}{2}  \bx_n \bx_n^{\top} \Weights  \bSigma^{-1} \Weights^{\top} \right]  + \text{tr}\left[ -\frac{1}{2}  \by_n \by_n^{\top}  \bSigma^{-1} \right] + \text{tr}\left[ \bx_n \by_n^{\top}  \bSigma^{-1} \Weights^{\top}  \right] \right)},
\end{split}
\end{align}
\normalsize
where $N$ is the number of data points. 
It can be seen that the natural parameters are $-\frac{1}{2}$, $-\frac{1}{2} \bx_n \bx_n^{\top}$, $-\frac{1}{2} \by_n \by_n^{\top}$ and $\bx_n \by_n^{\top}$, and the corresponding sufficient statistics are $\log |\bSigma|$, $\Weights \bSigma^{-1} \Weights^{\top}$, $\bSigma^{-1}$ and $\bSigma^{-1} \Weights^{\top}$.

\paragraph{Matrix-normal-inverse-Wishart posterior.}
Putting everything together, the natural parameters of the posterior 
\begin{equation*}
    p(\Weights, \bSigma | \bX, \bY) \propto  p(\bSigma; \bU_0, u_0) p(\Weights | \bSigma; \bM_0, \bR_0)  \prod_{n=1}^{N} p(\by_n | \bx_n, \bSigma, \Weights)
\end{equation*}
can now easily be computed using the above canonical forms of the prior and likelihood by simply taking the sum of the natural parameters. 
The resulting posterior natural parameters and corresponding sufficient statistics are summarised in Tab.~\ref{tab:posterior_nat_params_and_suff_stats}. 
\begin{table}[h!]
  \centering
  \caption{Natural parameters and sufficient statistics of the matrix-normal-inverse-Wishart prior and posterior.}
  \begin{tabular}{lll}
    \hline
    Prior natural parameters $\btheta_0$ & Posterior natural parameters $\btheta$ & Sufficient statistics \\
    \hline
    $\btheta_0^{(1)}= -\frac{1}{2}\bR_0^{-1}$ & $\btheta^{(1)}= -\frac{1}{2}(\bX^{\top} \bX + \bR_0^{-1})$ & $\Weights \bSigma^{-1} \Weights^{\top}$ \\
    $\btheta_0^{(2)}= \bR_0^{-1} \bM_0$ & $\btheta^{(2)}=\bX^{\top} \bY + \bR_0^{-1} \bM_0$ & $\bSigma^{-1} \Weights^{\top} $ \\
    $\btheta_0^{(3)}= -\frac{1}{2}(\bU_0 + \bM_0^{\top} \bR_0^{-1} \bM_0)$ & $\btheta^{(3)}=-\frac{1}{2}(\bY^{\top} \bY + \bU_0 + \bM_0^{\top} \bR_0^{-1} \bM_0)$ & $\bSigma^{-1}$ \\
    $\btheta_0^{(4)}= -\frac{1}{2}(u_o + D_{\by} + D_{\bx} + 1)$ & $\btheta^{(4)}=-\frac{1}{2}(N + u_o + D_{\by} + D_{\bx} + 1)$ & $\log |\bSigma|$ \\
    \hline
  \end{tabular}
  \label{tab:posterior_nat_params_and_suff_stats}
\end{table}
Transforming these back to the distribution parameters of the matrix-normal-inverse-Wishart distribution yields the parameters in \eqref{eq:method:layerwise_posterior_parameters}.

\paragraph{From natural parameters back to distribution parameters.}
Finally, we prove \eqref{eq:method:layerwise_posterior_parameters}. 
The first column (prior natural parameters) of Tab.~\ref{tab:posterior_nat_params_and_suff_stats} shows the \emph{forward parameter mapping} for the matrix-normal-inverse-Wishart prior, i.e.\ mapping from distribution parameters to natural parameters. 
The \emph{inverse parameter mapping}---i.e. mapping from natural parameters to distribution parameters---is straightforward to see, going through the natural parameters in the table row by row: $\bR_0 = -2 \btheta_0^{(1)}$, $\bM_0 = \bR_0 \btheta_0^{(2)}$, $\bU_0 = -2 \btheta_0^{(3)}$ and $u_0 = - 2 \btheta_0^{(4)} - (D_{\by} + D_{\bx} + 1)$. 
The inverse mapping for the posterior is of course the same, since it is from the same exponential family. 
Thus, the posterior parameters are given by
\begin{subequations}
\begin{align}
\bR &= \left(-2 \btheta^{(1)} \right)^{-1} = (\bX^{\top} \bX + \bR_0^{-1})^{-1}, \\
\bM &= \bR \btheta^{(2)} = (\bX^{\top} \bX + \bR_0^{-1})^{-1} (\bX^{\top} \bY + \bR_0^{-1} \bM_0), \\
\bU &= -2 \btheta^{(3)} - \bM^{\top} \bR^{-1} \bM = \bY^{\top} \bY + \bU_0 + \bM_0^{\top} \bR_0^{-1} \bM_0 - \bM^{\top} \bR^{-1} \bM, \\
u &= - 2 \btheta^{(4)} - (D_{\by} + D_{\bx} + 1) = N + u_0.
\end{align}
\end{subequations}

%% file: sec/appendix/complexity.tex
\section{Computational complexity}\label{app:computational_complexity}
The forward and backward pass to compute the pseudo-targets has the same computational complexity as a gradient-update step in standard gradient-based methods. 
Per layer, the complexity of a forward and backward pass is $O(B D_{\bx}  D_{\by})$ due to the matrix multiplication of the weights $\Weights \in \mathbb{R}^{D_{\bx} \times D_{\by}}$ and a mini-batch of size $\bX \in \mathbb{R}^{B \times D_{\bx}}$. 
The additional steps in BALI include updating the EMA of the natural parameters, computing the distribution parameters and sampling the weights: 
\begin{itemize}
\item The complexity of all EMA updates is of the order $O(D_{\bx}^2 B + D_{\bx} D_{\by} B + D^2_{\by} B)$, where each term corresponds to one of the $3$ matrix multiplications in \eqref{eq:method:ema_statistics}. 
\item Computing distribution parameters in \eqref{eq:method:layerwise_posterior_parameters} has complexity $O(D_{\bx}^3 + D_{\bx}^2 D_{\by} + D_{\bx} D^2_{\by} )$. 
The first term is due to the inverse of $\bR^{-1}$, which we compute via the Cholesky decomposition with complexity $O(D_{\bx}^3)$. 
The second and third term is due to the matrix multiplications, that is, computing $\bM$ has complexity $O(D_{\bx}^2 D_{\by})$ and $\bU$ has complexity $O(D_{\bx}^2 D_{\by} + D_{\bx} D^2_{\by})$. 
\item Sampling the weights via \eqref{eq:weight_sample_matrix_normal} requires sampling $D_{\bx} \cdot D_{\by}$ times from the standard normal to obtain the matrix $\bA$, leading to $O(D_{\bx} D_{\by})$. Sampling further requires computing the Cholesky decomposition of $\bR$ and $\bSigma$. The former can be obtained easily by inverting the already computed Cholesky factor of $\bR^{-1}$ in $O(D_{\bx}^2)$, i.e.\ without additional complexity. 
Computing the Cholesky of $\bU$ has complexity $O(D_{\by}^3)$. 
\end{itemize}
The overall complexity therefore is $O\big((D_{\bx}^2 + D_{\bx} D_{\by} + D^2_{\by}) B  + D_{\bx}^3 + D_{\by}^3 + D_{\bx}^2 D_{\by} + D_{\bx} D^2_{\by}  \big)$.
Therefore, when the batch size $B$ is similar or larger than the network dimensions, the complexity is of the same order as the forward pass and backward pass. 
On parallel hardware, the matrix inversion and sampling from the standard normal can be the dominating factor despite similar complexity. 
However, there are multiple properties that could be exploited for a fast implementation. 
For instance, sampling the independent normals and reshaping them into the matrix $\bA$ in \eqref{eq:weight_sample_matrix_normal} is independent from the rest of the algorithm and could be done at any time in parallel. 
Furthermore, the matrices $\bR^{-1}$ in each layer only depend on the input and could be inverted in parallel while the backward pass is computed. 
Similarly, the Cholesky decomposition of the covariance matrix $\bU$ of the top layers could already be computed in parallel while the gradients for the lower layers are computed. 
However, achieving computational efficiency was not in the scope of this paper and the experiments in this paper were performed on a single GPU. 
We point to several previous works that have achieved significant acceleration for KFAC-based methods (which shares many similarities and computational aspects with BALI), using an efficient distributed setup, see e.g.\ \cite{Ba2017Distributed,Shi2021Distributed, Zhang2022Scalable}.

%% file: sec/appendix/algorithm_comparison.tex
\section{BALI algorithm and comparison to other inference algorithms}
The pseudo-code for BALI is shown in \cref{alg:layerwise_mvblr}.
For reference, we additionally show the inference algorithms Noisy Adam, Vadam, VOGN and Noisy K-FAC, with the deterministic optimizer Adam and K-FAC, respectively, all in a similar notation for better comparability.
\begin{algorithm}[htb]
	\caption{BALI. For simplicity, we demonstrate the algorithm with batch size $B=1$.}
	\begin{algorithmic}[]\label{alg:layerwise_mvblr}
		\REQUIRE $\alpha$: stepsize
		\REQUIRE $\beta$: exponential moving average parameter
		\REQUIRE $N$: number of training examples
		\REQUIRE $\{\bM_{0,l}, \bR_{0,l}, \bU_{0,l}, \rmu_{0,l}\}_{l=1}^L$: prior parameters
		\STATE Initialise iteration step ${t \leftarrow 0}$
		\STATE Initialise EMA ${\{\bPsi^\mathrm{xx}_l, \bPsi^\mathrm{xy}_l, \bPsi^\mathrm{yy}_l, \bPsi^\mathrm{gg}_l\}_{l=1}^L \leftarrow 0}$ 
		\STATE Initialise weights $\{\Weights_l\}_{l=1}^L$
		\WHILE{stopping criterion not met}
		\STATE $t \leftarrow t + 1$
		\STATE $b \leftarrow 1 - (1 - \beta)^t$
		\STATE $\Big\{\mathbf{g}_l \leftarrow \nabla_{\mathbf{z}_l} \log p\big(\mathbf{y} | \mathbf{x}, \{\Weights_l\}_{l=1}^L\big)\Big\}_{l=1}^L$
		\FOR{each layer $l$}
		\STATE // \textit{Compute targets using norm.\ gradients (Sec.~\ref{sec:method:targets})}
		\STATE $\bPsi^\mathrm{gg}_l \leftarrow (1 - \beta) \cdot \bPsi^\mathrm{gg}_l + \beta \cdot \mathbf{g}_l^2$
		\STATE $\by_l \leftarrow \bz_l + \alpha \cdot \frac{\mathbf{g}_l}{\sqrt{\nicefrac{\bPsi^\mathrm{gg}_l}{b}}}$ 
		\\
		// \textit{Update exponential moving average (Sec.~\ref{sec:method:adaptation})}
		\STATE $\bPsi^\mathrm{xx}_l \leftarrow (1 - \beta) \cdot \bPsi^\mathrm{xx}_l + \beta \cdot N \cdot\bx_l\,\bx_l^\top$ 
		\STATE $\bPsi^\mathrm{xy}_l \leftarrow (1 - \beta) \cdot \bPsi^\mathrm{xy}_l + \beta \cdot N \cdot \bx_l\,\by_l^\top$
		\STATE $\bPsi^\mathrm{yy}_l \leftarrow (1 - \beta) \cdot \bPsi^\mathrm{yy}_l + \beta \cdot N \cdot\by_l\,\by_l^\top$ 
		\\
		// \textit{Compute distribution parameters (Sec.~\ref{sec:method:inference})} 
		\STATE $\bR_l \leftarrow \left( \bR_{0, l}^{-1} + \frac{\bPsi^\mathrm{xx}_l}{b} \right)^{-1}$ 
		\STATE $\bM_l \leftarrow \bR_l \left( \bR_{0,l}^{-1} \bM_{0,l}  + \frac{\bPsi^\mathrm{xy}_l}{b} \right)$
		\STATE $\bU_l \leftarrow \bU_{0, l} + \bM_{0, l}^\top \bR_{0, l}^{-1} \bM_{0, l} - \bM_l^\top \bR_l^{-1} \bM_l + \frac{\bPsi^\mathrm{yy}_l}{b}$
		\\
		// \textit{Estimate $\bSigma_l$ and sample weights (Sec.~\ref{sec:method:targets})} 
		\STATE $\bSigma_l \leftarrow \frac{1}{\rmu_{0,l} + N + D_{\by_l} + 1} \cdot \bU_l$ 
		\STATE $\Weights_l \sim \mathcal{MN}\left(\bM_l, \bR_l, \bSigma_l \right)$ 
		\ENDFOR
		\ENDWHILE
	\end{algorithmic}
\end{algorithm}

\begin{algorithm}[H]
\caption{Noisy Adam. Differences from standard Adam are shown in \textcolor{blue}{blue}.}
\begin{algorithmic}[1]\label{alg:noisy_adam}
    \REQUIRE $\alpha$: Stepsize
    \REQUIRE $\beta_1, \beta_2$: Exponential decay rates for updating $\mu$ and the Fisher $\mathbf{F}\approx\mathop{\mathrm{diag}}(\mathbf{f}) $
    \REQUIRE $\lambda, N, \eta, \gamma_{\text{ex}}$: KL weighting, number of training examples, prior variance, extrinsic damping term
    \STATE $\boldsymbol{\mu} \leftarrow \boldsymbol{\mu}_0, \mathbf{f} \leftarrow \mathbf{f}_0, \mathbf{m} \leftarrow 0, t\leftarrow 0$
    \STATE Calculate the intrinsic damping term $\textcolor{blue}{\gamma_{\text{in}} = \frac{\lambda}{N\eta}}$
    \WHILE{stopping criterion not met}
        \STATE $t \leftarrow t + 1$
        \STATE $\textcolor{blue}{\mathbf{w} \sim \mathcal{N}(\boldsymbol{\mu}, \frac{\lambda}{N} \text{diag}(\mathbf{f} + \gamma_{\text{in}})^{-1})}$
        \STATE $\mathbf{v} \leftarrow \nabla_\mathbf{w} \log p(y| \mathbf{x}, \mathbf{w}) \textcolor{blue}{- \gamma_{\text{in}} \cdot \mathbf{w}}$
        \STATE $\mathbf{m} \leftarrow \beta_1 \cdot \mathbf{m} + (1 - \beta_1) \cdot \mathbf{v}$ \COMMENT{Update momentum}
        \STATE $\mathbf{f} \leftarrow \beta_2 \cdot \mathbf{f} + (1 - \beta_2) \cdot(\nabla_\mathbf{w} \log p(y| \mathbf{x}, \mathbf{w}))^2$
        \STATE $\hat{\mathbf{m}} \leftarrow \frac{\mathbf{m}}{1 - \beta_1^t}$ \COMMENT{Adjust momentum}
        \STATE $\boldsymbol{\mu} \leftarrow \boldsymbol{\mu} + \alpha \cdot \frac{\hat{\mathbf{m}}}{\textcolor{blue}{\mathbf{f} + \gamma_{\text{in}} + \gamma_{\text{ex}}}}$ \COMMENT{Update parameters}
    \ENDWHILE
\end{algorithmic}
\end{algorithm}
\begin{algorithm}[H]
\caption{Vadam. Differences from standard Adam are shown in \textcolor{blue}{blue}.}
\begin{algorithmic}[1]\label{alg:vadam}
    \REQUIRE $\alpha$: Stepsize
    \REQUIRE $\beta_1, \beta_2$: Exponential decay rates for updating $\mu$ and the Fisher $\mathbf{F}\approx\mathop{\mathrm{diag}}(\mathbf{f}) $
    \REQUIRE $\lambda, N$: KL weighting, number of training examples
    \STATE $\boldsymbol{\mu} \leftarrow \boldsymbol{\mu}_0, \mathbf{f} \leftarrow \mathbf{f}_0, \mathbf{m} \leftarrow 0, t\leftarrow 0$
    \STATE Calculate the intrinsic damping term $\textcolor{blue}{\gamma_{\text{in}} = \frac{\lambda}{N}}$
    \WHILE{stopping criterion not met}
        \STATE $t \leftarrow t + 1$
        \STATE $\textcolor{blue}{\mathbf{w} \sim \mathcal{N}(\boldsymbol{\mu}, \frac{1}{N} \text{diag}(\mathbf{f} + \gamma_{\text{in}})^{-1})}$
        \STATE $\mathbf{v} \leftarrow \nabla_\mathbf{w} \log p(y| \mathbf{x}, \mathbf{w}) \textcolor{blue}{- \gamma_{\text{in}} \cdot \boldsymbol{\mu}}$
        \STATE $\mathbf{m} \leftarrow \beta_1 \cdot \mathbf{m} + (1 - \beta_1) \cdot \mathbf{v}$ \COMMENT{Update momentum}
        \STATE $\mathbf{f} \leftarrow \beta_2 \cdot \mathbf{f} + (1 - \beta_2) \cdot(\nabla_\mathbf{w} \log p(y| \mathbf{x}, \mathbf{w}))^2$
        \STATE $\hat{\mathbf{m}} \leftarrow \frac{\mathbf{m}}{1 - \beta_1^t}$ \COMMENT{Adjust momentum}
        \STATE $\hat{\mathbf{f}} \leftarrow \frac{\mathbf{f}}{1 - \beta_2^t}$
        \STATE $\boldsymbol{\mu} \leftarrow \boldsymbol{\mu} + \alpha \cdot \frac{\hat{\mathbf{m}}}{\sqrt{\hat{\mathbf{f}}} + \textcolor{blue}{\gamma_{\text{in}}}}$ \COMMENT{Update parameters}
    \ENDWHILE
\end{algorithmic}
\end{algorithm}
\begin{algorithm}[H]
\caption{VOGN. Differences from standard Adam are shown in \textcolor{blue}{blue}.}
\begin{algorithmic}[1]\label{alg:vogn}
    \REQUIRE $\alpha$: Stepsize
    \REQUIRE $\beta_1, \beta_2$: Exponential decay rates for updating $\mu$ and the Fisher $\mathbf{F}\approx\mathop{\mathrm{diag}}(\mathbf{f}) $
    \REQUIRE $\lambda_0, \lambda_T, T_\lambda$: initial KL weighting, final KL weighting, KL annealing steps
    \REQUIRE $N, \eta, \gamma_{\text{ex}}$: number of training examples, prior variance, extrinsic damping term
    \STATE $\boldsymbol{\mu} \leftarrow \boldsymbol{\mu}_0, \lambda \leftarrow \lambda_0, \mathbf{f} \leftarrow \mathbf{f}_0, \mathbf{m} \leftarrow 0, t\leftarrow 0$
    \STATE Calculate the intrinsic damping term $\textcolor{blue}{\gamma_{\text{in}}(\lambda) = \frac{\lambda}{N\eta}}$
    \WHILE{stopping criterion not met}
        \STATE $t \leftarrow t + 1$
        \STATE $\textcolor{blue}{\mathbf{w} \sim \mathcal{N}(\boldsymbol{\mu}, \frac{1}{N} \text{diag}(\mathbf{f} + \gamma_{\text{in}}(\lambda) + \gamma_{\text{ex}})^{-1})}$
        \STATE $\mathbf{v} \leftarrow \nabla_\mathbf{w} \log p(y| \mathbf{x}, \mathbf{w}) \textcolor{blue}{- \gamma_{\text{in}}(\lambda) \cdot \boldsymbol{\mu}}$
        \STATE $\mathbf{m} \leftarrow \beta_1 \cdot \mathbf{m} + \textcolor{blue}{1} \cdot \mathbf{v}$ \COMMENT{Update momentum}
        \STATE $\mathbf{f} \leftarrow \textcolor{blue}{(1 - \lambda\,\beta_2)}  \cdot \mathbf{f} + \textcolor{blue}{\beta_2} \cdot(\nabla_\mathbf{w} \log p(y| \mathbf{x}, \mathbf{w}))^2$
        \STATE $\boldsymbol{\mu} \leftarrow \boldsymbol{\mu} + \alpha \cdot \frac{\textcolor{blue}{\mathbf{m}}}{\textcolor{blue}{\mathbf{f} + \gamma_{\text{in}}(\lambda) + \gamma_{\text{ex}}}}$ \COMMENT{Update parameters}
        \STATE $\textcolor{blue}{\lambda \leftarrow \text{LinearSchedule}(\lambda, t; \lambda_T, T_\lambda)}$ \COMMENT{Update KL weigthing}
    \ENDWHILE
\end{algorithmic}
\end{algorithm}
\begin{algorithm}[H]
\caption{Noisy K-FAC. Note that the authors assume zero momentum for simplicity. Differences from standard K-FAC
are shown in \textcolor{blue}{blue}.}
\begin{algorithmic}[1]\label{alg:noisy_kfac}
    \REQUIRE $\alpha$: stepsize
    \REQUIRE $\beta$: exponential moving average parameter
    \REQUIRE $\lambda, N, \eta$: KL weighting, number of training examples, prior variance
    \STATE $\{\bPsi_l^\textrm{xx}\}_{l=1}^L \leftarrow 0, \{\bPsi_l^\textrm{gg}\}_{l=1}^L \leftarrow 0, t \leftarrow 0$
    \STATE Initialise $\{\Weights_l\}_{l=1}^L$
    \STATE Compute the intrinsic damping term \textcolor{blue}{$\gamma_{\text{in}} = \frac{\lambda}{N\eta}$}
    \WHILE{stopping criterion not met}
        \STATE $t \leftarrow t + 1$
        \STATE $\Big\{\mathbf{g}_l \leftarrow \nabla_{\mathbf{z}_l} \log p\big(\mathbf{y} | \mathbf{x}, \{\Weights_l\}_{l=1}^L\big)\Big\}_{l=1}^L$ \COMMENT{Compute gradients w.r.t.\ layer outputs}
        \FOR{each layer $l$}
            \STATE $\bPsi_l^\textrm{xx} \leftarrow (1-\beta) \cdot \bPsi_l^\textrm{xx} + \beta \cdot \mathbf{x}_l\,\mathbf{x}_l^\top$
            \COMMENT{Update exponential moving average}
            \STATE $\bPsi_l^\textrm{gg} \leftarrow (1-\beta) \cdot \bPsi_l^\textrm{gg} + \beta \cdot \mathbf{g}_{l}\,\mathbf{g}_{l}^\top$
            \STATE $\bR_l \leftarrow \left(\pi_l \sqrt{\textcolor{blue}{\gamma_{\text{in}}}}\,\mathbf{I} + \bPsi_l^\textrm{xx}\right)^{-1}$
            \COMMENT{Compute first scale}
            \STATE $\bSigma_l \leftarrow \left(\frac{1}{\pi_{l}} \sqrt{\textcolor{blue}{\gamma_{\text{in}}}}\,\mathbf{I} + \bPsi_l^\textrm{gg}\right)^{-1}$
            \COMMENT{Compute second scale}
            \STATE $\mathbf{G}_l \leftarrow \mathbf{x}_l\,\mathbf{g}_{l}^\top \textcolor{blue}{- \gamma_{\text{in}} \cdot \Weights_l}$
            \COMMENT{Compute gradients w.r.t.\ weights}
            \STATE $\bM_l \leftarrow \bM_l + \alpha \cdot \bR_l\,\mathbf{G}_l\,\bSigma_l$
            \COMMENT{Compute location}
            \STATE \textcolor{blue}{$\Weights_l \sim \mathcal{MN}\left(\bM_l, \bR_l, \bSigma_l \right)$}
            \COMMENT{Sample weight matrix from posterior}
        \ENDFOR
    \ENDWHILE
\end{algorithmic}
\end{algorithm}

%% file: sec/appendix/mean_updates.tex
\section{Noise reparametrisation}\label{app:noise_reparam}
In \eqref{eq:methods:pseudo_targets} of Sec.~\ref{sec:method:targets}, we defined the pseudo-targets as the gradient-updated layer outputs. 
In this formulation, the layer outputs are obtained via the forward pass with a randomly drawn weight from the previous posterior approximation. 
This can be seen as analogous to standard \emph{weight reparametrisation} in Bayesian neural networks. 
However, we can also sample from the predictive distribution over the (linear) layer outputs induced by the previous posterior approximation, which corresponds to \emph{local reparametrisation}. 

To show this, we define the pseudo-targets in a slightly different (but equivalent) form that more clearly differentiates between these types of sampling. 
Furthermore, we use this formulation in App.~\ref{app:mean_updates} to derive the mean updates that result from the EMA updates of the natural parameters.
To this end, we first write the pseudo-targets from \eqref{eq:methods:pseudo_targets} in terms of the posterior mean and the sampling noise as
\begin{align}\label{eq:pseudo_targets_redefinition}
\begin{split}
\by_{n,t} 
&= \bz_{n,t} + \alpha \cdot \frac{\mathbf{g}_{n,t}}{\sqrt{\nicefrac{\bPsi^{\mathrm{gg}}_t}{b_t}}} \\
&= \Weights_{t}^{\top} \bx_{n,t} + \alpha \cdot \frac{\mathbf{g}_{n,t}}{\sqrt{\nicefrac{\bPsi^{\mathrm{gg}}_t}{b_t}}} \\
&= \bM_{t-1}^{\top} \bx_{n,t} + \boldsymbol{\epsilon}_{n,t} + \alpha \cdot \frac{\mathbf{g}_{n,t}}{\sqrt{\nicefrac{\bPsi^{\mathrm{gg}}_t}{b_t}}},
\end{split}
\end{align}
where $\boldsymbol{\epsilon}_{n,t} \in \mathbb{R}^{D_{\by}}$ is the noise induced at the output nodes due to sampling the weights. 
More generally, we now redefine the pseudo-targets via the forward pass using the \emph{mean} of the previous posterior approximation:
\begin{align}
\by_{n,t} &= \bM_{t-1}^{\top} \bx_{n,t} + \Delta \by_{n,t}, ~~~~~ \Delta \by_{n,t} := \boldsymbol{\epsilon}_{n,t} + \alpha \cdot \frac{\mathbf{g}_{n,t}}{\sqrt{\nicefrac{\bPsi^\mathrm{gg}_{t}}{b_{t}}}},
\end{align}
where we now define the \emph{noise} $\boldsymbol{\epsilon}_{n,t}$ as part of the \emph{target update} $\Delta \by_{n,t}$.
For an entire batch of size $B$, we then have
\begin{align}\label{eq:pseudo_targets_redefinition_batched}
\bY_{t} = \bX_{t} \bM_{t-1} + \Delta \bY_{t}, 
~~~~~ \Delta \bY_{t} = \mathbf{E}_{t} + \alpha \cdot \frac{\mathbf{G}_{t}}{\sqrt{\nicefrac{\bPsi^\mathrm{gg}_{t}}{b_{t}}}},
\end{align}
where the noise is a matrix $\mathbf{E}_t \in \mathbb{R}^{B \times D_{\by}}$. 
Similarly, the gradients w.r.t.\ the noisy layer output nodes are a matrix $\mathbf{G}_t \in \mathbb{R}^{B \times D_{\by}}$ and the division with the square-root is element-wise, i.e.\ independent for all $B$ examples. 

\paragraph{Weight reparametrisation.}
In the above definition of the pseudo-targets, we defined $\mathbf{Z}_{t} = \bX_{t}^{\top} \Weights_{t}  = \bX_{t}^{\top} \bM_{t} + \mathbf{E}_{t}$, which of course implies that the output noise is $\bE_t = \bX_{t} (\Weights_{t} - \bM_{t-1})$.
And since
$
\Weights_{t} \sim \mathcal{MN}\left(\bM_{t-1}, \bR_{t-1}, \bSigma_{t-1} \right),
$
the noise distribution is given by
\begin{align}
\bE_t &\sim \mathcal{MN}\left( \mathbf{0}, \, \bX_{t} \bR_{t-1} \bX_{t}^{\top}, \, \bSigma_{t-1} \right).
\end{align}
We refer to sampling the weight noise as \emph{weight reparametrisation}.

\paragraph{Local reparametrisation.}
Instead of computing the linear transformation $\bz_{n,t} = \Weights_t^{\top} \bx_{n,t} \,$ using a \emph{weight sample}
\begin{align}
\Weights_t 
&\sim \mathcal{MN}\left(\bM_{t-1}, \bR_{t-1}, \bSigma_{t-1} \right),
\end{align}
we can compute the distribution induced by the linear transformation in closed-form. 
This distribution over output node values again follows a matrix-normal distribution with
\begin{align}\label{eq:local_reparam_predictive}
\Weights^{\top} \bx_{n,t} &\sim \mathcal{MN}\left( \bM_{t-1}^{\top} \bx_{n,t}, \, \bx_{n,t}^{\top} \bR_{t-1} \bx_{n,t}, \, \bSigma_{t-1} \right).
\end{align}
Since the predictive distribution in \eqref{eq:local_reparam_predictive} already has mean $\Weights_t^{\top} \bx_{n,t}$ in accordance with the redefinition of the pseudo-targets in \eqref{eq:pseudo_targets_redefinition}, we see that the noise on the output nodes is distributed as
\begin{align}\label{eq:local_reparam_noise_dist}
\boldsymbol{\epsilon}_{n,t} &\sim \mathcal{MN}\left( \mathbf{0}, \, \bx_{n,t}^{\top} \bR_{t-1} \bx_{n,t}, \, \bSigma_{t-1} \right).
\end{align}
We refer to sampling from the transformed distribution as \emph{local reparametrisation}, since it is analogous to the so-called local reparametrisation trick \cite{Kingma2015LocalReparam}.
Sampling from the distribution in \eqref{eq:local_reparam_noise_dist} requires computing the Cholesky decomposition of the two scale-matrices. 
For the scale-matrix $\bx_{n,t}^{\top} \bR_{t-1} \bx_{n,t}$, this is just a square-root of the resulting scalar, because we ignore the correlation between different examples $\bx_{n,t}$. 
As a consequence, the noise variance is reduced compared to the \emph{weight reparametrisation}.

\paragraph{Deterministic transformation.}
A third alternative is to ignore the sampling noise in the computation of the pseudo-targets and the natural parameters. 
In this case, we simply have
\begin{align}
\bY_{t} = \bX_{t} \bM_{t-1} + \Delta \bY_{t}, 
~~~~~ \Delta \bY_{t} = \alpha \cdot \frac{\mathbf{G}_{t}}{\sqrt{\nicefrac{\bPsi^\mathrm{gg}_{t}}{b_{t}}}}.
\end{align}
This approach ignores the noise due to the epistemic uncertainty while learning the neural network.

\section{Posterior mean updates}\label{app:mean_updates}
As shown in App.~\ref{sec:appendix:MVBLR}, the posterior means are given by \eqref{eq:method:layerwise_posterior_mean}. 
However, as described in Sec.~\ref{sec:method:adaptation}, we use an EMA to update the natural parameters stemming from the likelihood, as defined by \eqref{eq:method:ema_statistics}.
Here we derive a simple expression for the mean update, resulting from the EMA update of the natural parameters. 
To achieve this, we will use the pseudo-target definition $\bY_{t} = \bX_{t} \bM_{t-1} + \Delta \bY_{t}$ from App.~\ref{app:noise_reparam} in order to express the new mean $\bM_t$ as a function of $\bM_{t-1}$.

Assume that we have already updated $\bPsi^{\mathrm{xx}}_t$ at iteration $t$ and inverted the precision matrix to compute the posterior scale matrix $\bR_t = \big(\bPsi^{\mathrm{xx}}_t + \bR_0^{-1}\big)^{-1}$. 
Then, we can write the mean at iteration $t$ as
\begin{align}\label{eq:app:mean_update_initial}
    \bM_t = \bR_t (\bPsi^{\mathrm{xy}}_t + \bR_0^{-1} \bM_0).
\end{align}
Inserting the EMA estimate for $\bPsi^{\mathrm{xy}}_t$ from \eqref{eq:method:ema_statistics} into \eqref{eq:app:mean_update_initial} gives
\begin{align}\label{eq:app:mean_update_via_precision_mean_ema}
\bM_t 
= \bR_t \left[ (1-\beta) \cdot \bPsi^{\mathrm{xy}}_{t-1} + \beta \cdot \frac{N}{B} \bX_t^{\top} \bY_t + \bR_0^{-1} \bM_0 \right].
\end{align}
Next, using \eqref{eq:app:mean_update_initial}, we can trivially express the natural parameter estimate of the previous iteration $t-1$ as
\begin{align}\label{eq:app:prev_precision_mean}
	\bPsi^{\mathrm{xy}}_{t-1} = \bR^{-1}_{t-1} \bM_{t-1} - \bR_0^{-1} \bM_0.
\end{align}
Furthermore, we expand the precision-mean update term $\bX_t^{\top} \bY_t$ by inserting the pseudo-target definition from App.~\ref{app:noise_reparam}: 
\begin{align}\label{eq:app:precision_mean_via_target_diff}
\bX_t^{\top} \bY_t 
&= \bX_t^{\top} \left( \bX_t \bM_{t-1} + \Delta \bY_t  \right) 
= \bX_t^{\top} \bX_{t} \bM_{t-1} + \bX_t^{\top} \Delta \bY_t.
\end{align}
Plugging \eqref{eq:app:prev_precision_mean} and \eqref{eq:app:precision_mean_via_target_diff} into \eqref{eq:app:mean_update_via_precision_mean_ema}, we obtain
\begin{align} \label{eq:app:mean_update}
\begin{split}
\bM_t 
&= \bR_t \left[ (1-\beta) \cdot (\bR^{-1}_{t-1} \bM_{t-1} - \bR_0^{-1} \bM_0) + \beta \cdot \frac{N}{B} \cdot (\bX_t^{\top} \bX_{t} \bM_{t-1} + \bX^{\top} \Delta \bY_t) + \bR_0^{-1} \bM_0 \right] \\
&= \bR_t \left[ (1-\beta) \cdot \bR^{-1}_{t-1}  + \beta \cdot \frac{N}{B} \cdot \bX_t^{\top} \bX_{t} \right] \bM_{t-1} +  \beta \cdot \frac{N}{B} \cdot \bR_t \bX_t^{\top} \Delta \bY_t + \beta \cdot \bR_t \bR_0^{-1} \bM_0 \\
&= \bR_t \left[ \bR^{-1}_{t} - \beta \cdot \bR_0^{-1}\right] \bM_{t-1} +  \beta \cdot \frac{N}{B} \cdot \bR_t \bX_t^{\top} \Delta \bY_t + \beta \cdot \bR_t \bR_0^{-1} \bM_0 \\
&= \bM_{t-1} + \beta \cdot \bR_t \Big( \bR_0^{-1} (\bM_0 - \bM_{t-1}) + \frac{N}{B} \cdot \bX_t^{\top} \Delta \bY_t \Big).
\end{split}
\end{align}
where the second line follows from the EMA update equations for $\bPsi^{\mathrm{xx}}_{t}$ and $\bR_{t-1}^{-1} = \bPsi^{\mathrm{xx}}_{t-1} + \bR_0^{-1}$. 
Hence, we can use \eqref{eq:app:mean_update} as an alternative implementation for parametrising and updating the means directly. 
Note again that we have used a rather general formulation, where we defined the pseudo-targets using a deviation $\Delta \bY$ from the posterior-predictive mean $\bX \bM$. 
Hence, these mean updates are applicable to pseudo-targets obtained via weight reparametrisation (as used in this paper), local reparametrisation and deterministic transformation (cf.~Sec.~\ref{app:noise_reparam}). 
Furthermore, it is applicable even if the pseudo-targets are obtained through other (potentially gradient-free) methods. 

%% file: sec/appendix/experiment_details.tex
\section{Experiment implementation details}\label{app:experiment_details}
\subsection{Synthetic data}\label{app:experiment_details:synthetic}
For the sinc-function experiment, we used the scaled and shifted sinc-function, $f(x) = b \cdot \frac{\sin (a x)}{a x} + c$, with $a=20, b=2, c=-1$ as the ground-truth function. 
We generated $256$ data points for training. 
The inputs were drawn random uniformly in the range $[-1, 1]$, and we applied additive Gaussian noise with a standard deviation of $0.1$ to the function outputs. 
For the sines-function experiment, we used the function $f(x) = a \cdot \sin(2 \pi x) + b\cdot \sin (4 \pi x) + c \cdot x$, where $a=0.3, b=0.3, c=1.0$.
We generated $32$ data points for training. 
The inputs were drawn random uniformly in two disconnected ranges $[-1.0, -0.25]$ and $[0.25, 1.0]$, and we applied additive Gaussian noise with a standard deviation of $0.02$ to the function outputs. 
For the two-moons function we generated 128 training data points and applied noise with a standard deviation of $0.2$.
For all 3 experiments, we used a neural network with $3$ hidden layers of $256$ units in each layer and a tanh activation function. 
In case of regression, the initial weights were drawn randomly from $\weights \sim \mathcal{N}(0, \sigma^2_{\text{init}} \cdot \mathbf{I})$ with $\sigma_{\text{init}} = 3$, the prior scale matrix term $\bR_0 = \sigma^{2}_r \cdot \mathbf{I}$ uses $\sigma_r = 80$ and $\bU_0 = \sigma^2_u \cdot \mathbf{I}$ uses $\sigma^2_u = N \cdot 0.01$. 
In case of the synthetic classification experiment, $\sigma_{\text{init}} = 1$, $\sigma_r = 20$, and $\sigma^2_u = N \cdot 0.01$.
The predictive distributions are shown as the mean and one standard deviation, computed using 256 samples of the weights.

\subsection{Regression benchmark}
We evaluated the performance of our model using $20$ random splits for each dataset, where $90 \%$ is used for training and $10 \%$ for testing. 
We used a neural network with a single hidden layer of 50 units and RELU activation function for all datasets except Protein, where we used 100 units as in previous work.
For regression tasks, we use the \emph{actual} targets in the last layer $L$ rather than pseudo-targets. 
Furthermore, we use the inferred (co-)variance $\bSigma_L$ to define a Gaussian likelihood and we obtain the pseudo-targets via gradients of the log-likelihood objective. 
For the initial forward pass before we have a posterior approximation, we draw the initial weights from $\weights_l \sim \mathcal{N}(0, \sigma^2_{\text{init}} \cdot \mathbf{I})$ with $\sigma_{\text{init}} = 1$. 
Other hyperparameters of BALI are summarised in Tab.~\ref{tab:classification_benchmark:hyperparameters}.
\begin{table}[htb!]
    \centering
    \caption{
Hyperparameters of BALI used for the regression benchmark in Sec.~\ref{sec:experiments:regression}, where $\alpha$ is the gradient step size and $\beta$ is the initial update rate, which we decay by a factor $5$ after $60\%$ of the total number of iterations and another factor $5$ after $80 \%$ of the total number of iterations. We define the prior covariance matrices as the diagonals $\bR_0 = \sigma_r^2 \cdot \mathbf{I}$ and $\bU_0 = \sigma_u^2 \cdot \mathbf{I}$, respectively. Furthermore, we use the effective dataset size $N_{\text{eff}}$ (instead of the actual dataset size $N$) for the EMA updates in \eqref{eq:method:ema_statistics} and estimate the natural parameters using the batch size $B$. 
}
    \label{tab:classification_benchmark:hyperparameters}
    \begin{tabular}{ll|ccccccc}
        \textbf{Dataset} & size & $\alpha$ & $\beta$ &  $\sigma^2_r$ & $\sigma^2_u$ & $N_{\text{eff}}$ & $B$ & $\#$iterations\\
        \midrule
        Yacht      & 308   & $0.3$ & $0.2$ & $40$ & $0.01\cdot N_{\text{eff}}$ & $N$   & 276 (=$N$) & 20000 \\
        Concrete   & 1030  & $0.3$ & $0.2$ & $40$ & $0.01\cdot N_{\text{eff}}$ & $N$   & 927 (=$N$) & 20000 \\
        Energy     & 768   & $0.3$ & $0.2$ & $40$ & $0.01\cdot N_{\text{eff}}$ & $N$   & 691 (=$N$) & 20000 \\
        Redwine    & 1599  & $0.3$ & $0.2$ & $40$ & $0.01\cdot N_{\text{eff}}$ & $N$   & 1024       & 20000 \\
        Kin8nm     & 8192  & $0.3$ & $0.2$ & $40$ & $0.01\cdot N_{\text{eff}}$ & $N/4$ & 1024       & 20000 \\
        Powerplant & 9568  & $0.3$ & $0.2$ & $40$ & $0.01\cdot N_{\text{eff}}$ & $N/4$ & 1024       & 20000 \\
        Protein    & 45730 & $0.3$ & $0.1$ & $20$ & $0.01\cdot N_{\text{eff}}$ & $N/4$ & 2048       & 30000 \\
        \bottomrule
    \end{tabular}
\end{table}

\subsection{Classification benchmark}
For the datasets Diabetes, Telescope and Spam from the classification benchmark, we used $5$ random splits for each dataset, where $80 \%$ is used for training and $20 \%$ for testing. 
For MNIST and FashionMNIST, we used the provided split with $60000$ and $10000$ samples for training and testing and we averaged the results obtained from $3$ random seeds. 
We learned each classification dataset with a neural network consisting of $2$ hidden layers of $256$ hidden units and "Leaky-Tanh" activations, i.e. $f(x) = \mathrm{tanh}(x) + 0.1 x$. 
For all experiments of the classification benchmark, we draw the initial weights (before we have our first layerwise posterior approximation) from $\weights_l \sim \mathcal{N}(0, \sigma^2_{\text{init}} \cdot \mathbf{I})$ with $\sigma_{\text{init}} = 1$. 
Other hyperparameters of BALI are summarised in Tab.~\ref{tab:classification_benchmark:hyperparameters}.
\begin{table}[h!]
    \centering
    \caption{Hyperparameters of BALI used for the classification benchmark in Sec.~\ref{sec:experiments:classification}, where $\alpha$ is the step size, $\beta$ is the initial update rate (decay by factor $5$ after $60\%$  and$80 \%$ of the total number of iterations). The prior covariances are again defined as $\bR_0 = \sigma_r^2 \cdot \mathbf{I}$ and $\bU_0 = \sigma_u^2 \cdot \mathbf{I}$, respectively. Furthermore, we use the effective dataset size $N_{\text{eff}}$ (instead of the actual dataset size $N$) for the EMA updates in \eqref{eq:method:ema_statistics} and estimate the natural parameters using the batch size $B$.  
}
    \label{tab:classification_benchmark:hyperparameters}
    \begin{tabular}{ll|ccccccc}
        \textbf{Dataset} & size & $\alpha$ & $\beta$ &  $\sigma^2_r$ & $\sigma^2_u$ & $N_{\text{eff}}$ & $B$ & $\#$iterations\\
        \midrule
        Spam       & 4601   & $0.1$ & $0.1$ & $10$ & $0.01\cdot N_{\text{eff}}$ & $\nicefrac{N}{4}$ & $2048$ & $20000$ \\
        Telescope  & 19020  & $0.5$ & $0.1$ & $8$ & $0.01\cdot N_{\text{eff}}$ & $\nicefrac{N}{10}$ & $2048$ & $20000$ \\
		Diabetes   & 253680 & $0.25$ & $0.05$ & $10$ & $0.01\cdot N_{\text{eff}}$ & $\nicefrac{N}{100}$ & $2048$ & $30000$ \\
        MNIST      & 70000  & $0.25$ & $0.1$ & $10$ & $0.01\cdot N_{\text{eff}}$ & $\nicefrac{N}{50}$ & $2048$ & $30000$ \\
        F-MNIST    & 70000  & $0.25$ & $0.1$ & $10$ & $0.01\cdot N_{\text{eff}}$ & $\nicefrac{N}{50}$ & $2048$ & $30000$ \\
        \bottomrule
    \end{tabular}
\end{table}

The hyperparameters of Monte Carlo DO, where weight are learned via the Adam optimiser are summarised in Tab.~\ref{tab:classification_benchmark:hyperparameters_MAPDO}. 
For the deterministic MAP estimation, we used the same hyper-parameters (except that the dropout probability is 0).
\begin{table}[htb!]
    \centering
    \caption{Hyperparameters of Monte Carlo DO used for the classification benchmark in Sec.~\ref{sec:experiments:classification}.}
    \label{tab:classification_benchmark:hyperparameters_MAPDO}
    \begin{tabular}{lccccc}
        \textbf{Dataset} & learning rate & weight decay & dropout probability & batch size $B$ & $\#$iterations \\
        \midrule
        Spam       & $0.001$ & $0.0001$ & $0.1$ & $2048$ & $20000$ \\
        Telescope  & $0.001$ & $0.0001$ & $0.1$ & $2048$ & $20000$ \\
		Diabetes   & $0.001$ & $0.0001$ & $0.1$ & $2048$ & $30000$ \\
        MNIST      & $0.002$ & $0.0001$ & $0.1$ & $2048$ & $30000$ \\
        F-MNIST    & $0.002$ & $0.0001$ & $0.1$ & $2048$ & $30000$ \\
        \bottomrule
    \end{tabular}
\end{table}

\section{Further results.}
\subsection{UCI regression benchmark log-likelihood}
Additionally to the RMSE in Tab.~\ref{tab:uci} of the main text, we provide the log-likelihood results for the UCI benchmark in Tab.~\ref{tab:results}.
\begin{table}[htb]
	\centering
	\caption{Averaged test log-likelihood for the regression benchmark (cf.~Sec.~\ref{sec:experiments:regression}). Higher is better.}
	\label{tab:results}
	\begin{adjustbox}{width=1.0\textwidth}
		\begin{tabular}{lccccccc}
			\toprule
			\textbf{Dataset} & \textbf{BALI} & \textbf{NNG-MVG} & \textbf{VADAM} & \textbf{BBB} & \textbf{PBP} & \textbf{DO} \\
			\midrule
			Yacht  & $\bold{-0.745 \pm 0.033}$ & $-2.316\pm0.006$ & $-1.70 \pm 0.03$ & $-2.408\pm0.007$ & $-1.634\pm0.016$ & $-1.250\pm0.015$ \\
			Concrete   & $\bold{-2.976 \pm 0.079}$ & $-3.039\pm0.025$ & $-3.39 \pm 0.02$ & $-3.149\pm0.018$ & $-3.161\pm0.019$ &  $\bold{-2.937\pm0.025}$ \\
			Energy     & $\bold{-0.610 \pm 0.035}$ & $-1.421\pm0.005$ & $-2.15 \pm 0.07$ & $-1.500\pm0.006$ & $-2.042\pm0.019$ &  $-1.212\pm0.005$ \\
			Redwine    & $-1.078 \pm 0.033$ & $-0.969\pm0.014$ & $-1.01 \pm 0.01$ & $-0.977\pm0.017$ & $-0.968\pm0.014$ &  $\bold{-0.928\pm0.013}$ \\
			Kin8nm     & $\phantom{+} \bold{1.175 \pm 0.008}$ &  $\phantom{+}1.148\pm0.007$ & $\phantom{+}0.76 \pm 0.00$ & $\phantom{+}1.111\pm0.007$ & $\phantom{+} 0.896\pm0.006$ &  $\phantom{+}1.136\pm0.007$ \\
			Powerplant & $\bold{-2.769 \pm 0.007}$ & $\bold{-2.776\pm0.011}$ & $-2.88 \pm 0.01$ & $-2.807\pm0.010$ & $-2.837\pm0.009$ &  $-2.808\pm0.009$ \\
			Protein & $\bold{-2.836 \pm 0.004}$ & $\bold{-2.836 \pm 0.002}$ & $ - $ & $-2.882 \pm 0.004$ & $- 2.896 \pm 0.004$ & $-2.871 \pm 0.001$ \\
			\bottomrule
		\end{tabular}
	\end{adjustbox}
\end{table}

\subsection{Bayes By Backprop trained on synthetic data}\label{sec:app:BBB}
\begin{figure}[h!]
  \centering
  \subfigure[KL weighting factor $1.0$.]{\includegraphics[width=0.325\textwidth]{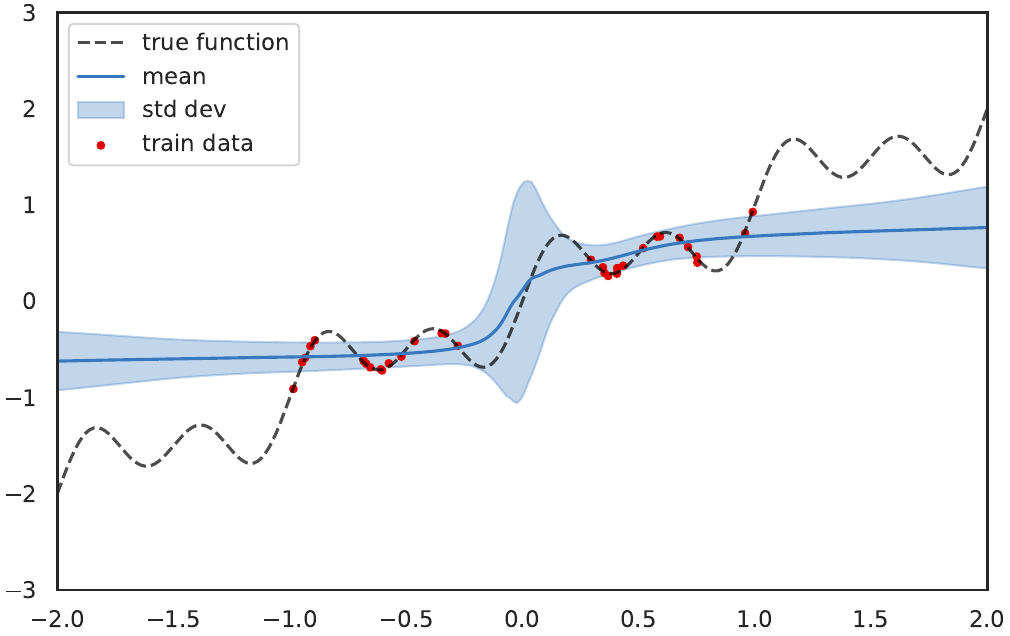}}
  \subfigure[KL weighting factor $0.1$.]{\includegraphics[width=0.325\textwidth]{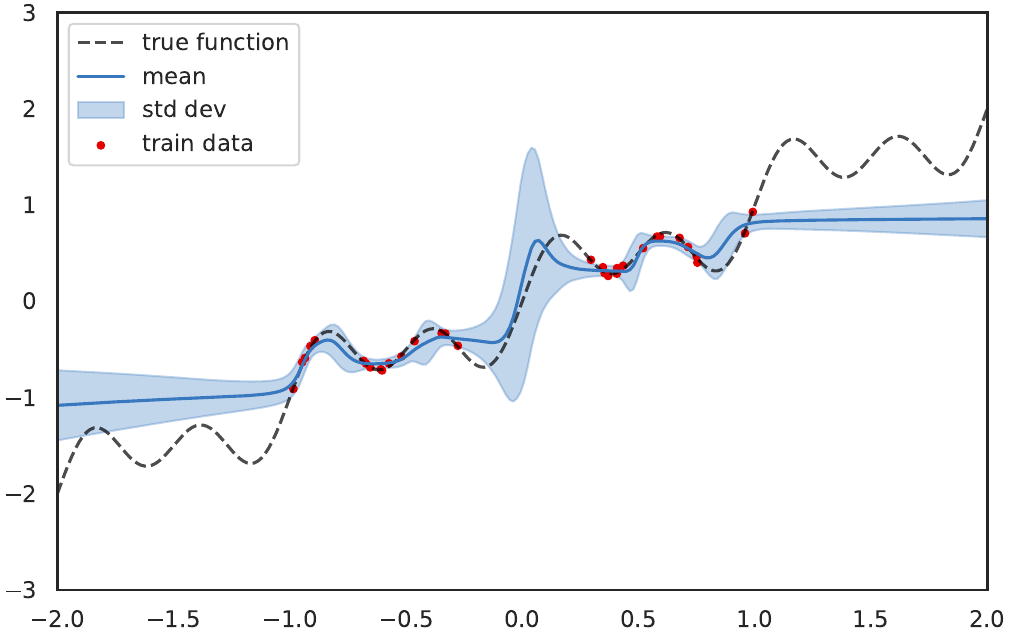}}
  \hfill
  \subfigure[KL weighting factor $0.01$.]{\includegraphics[width=0.325\textwidth]{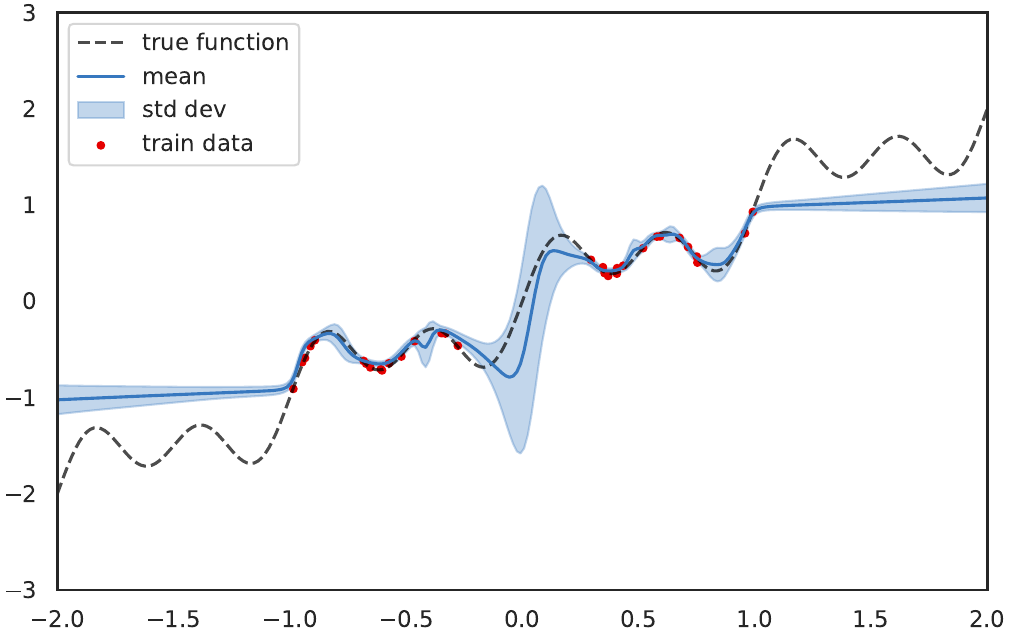}}
  \caption{Bayes By Backprop posterior-predictive distribution on the sines-trend dataset (cf.~Sec.~\ref{sec:experiments:toy}).}
  \label{fig:app:BBB_sine}
\end{figure}

\begin{figure}[h!]
  \centering
  \subfigure[KL weighting factor $1.0$.]{\includegraphics[width=0.325\textwidth]{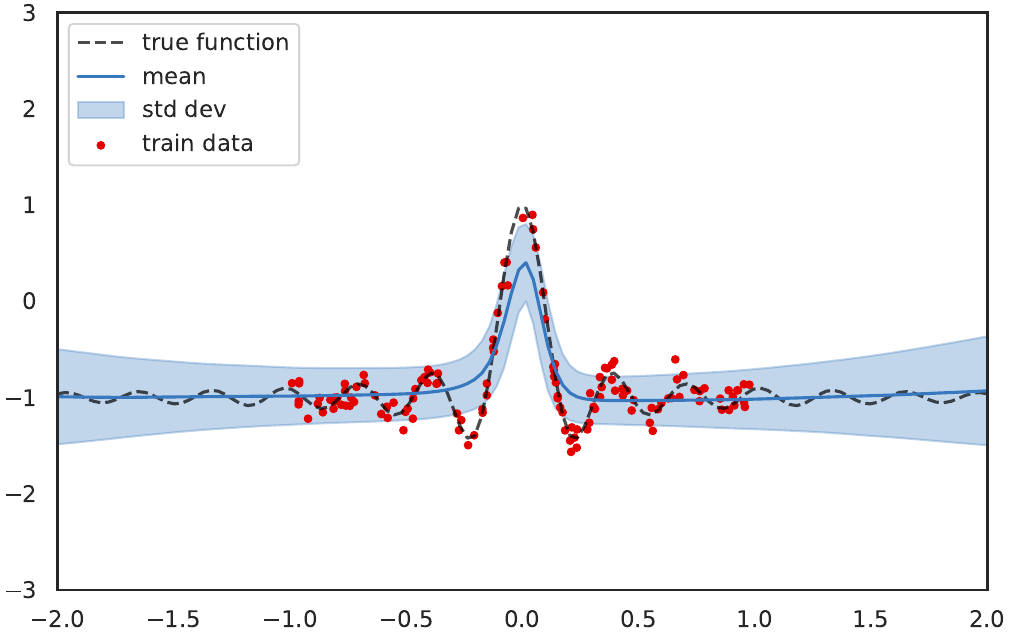}}
  \subfigure[KL weighting factor $0.1$.]{\includegraphics[width=0.325\textwidth]{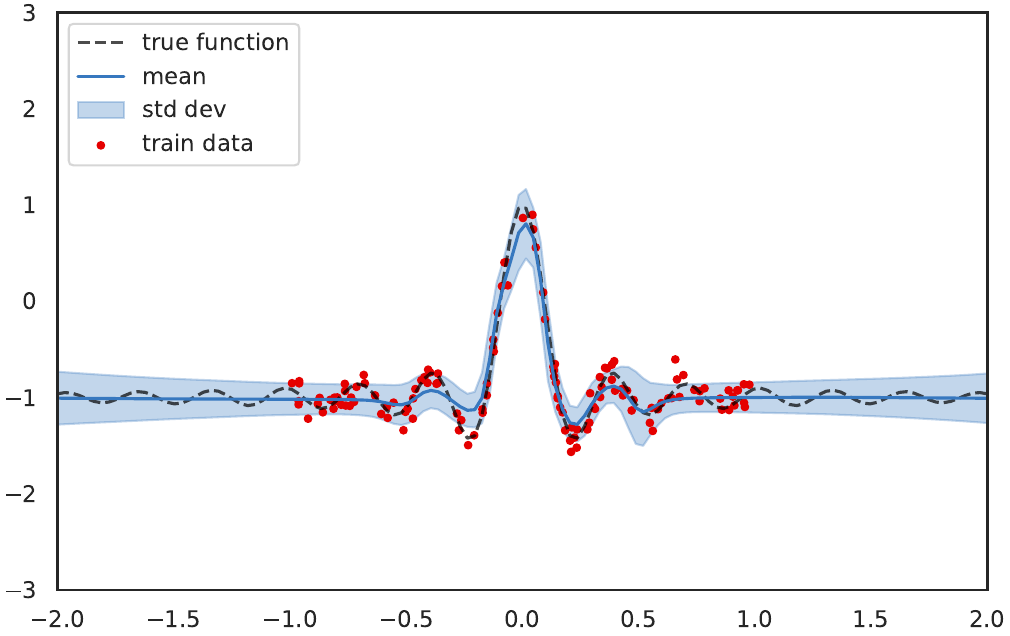}}
  \hfill
  \subfigure[KL weighting factor $0.01$.]{\includegraphics[width=0.325\textwidth]{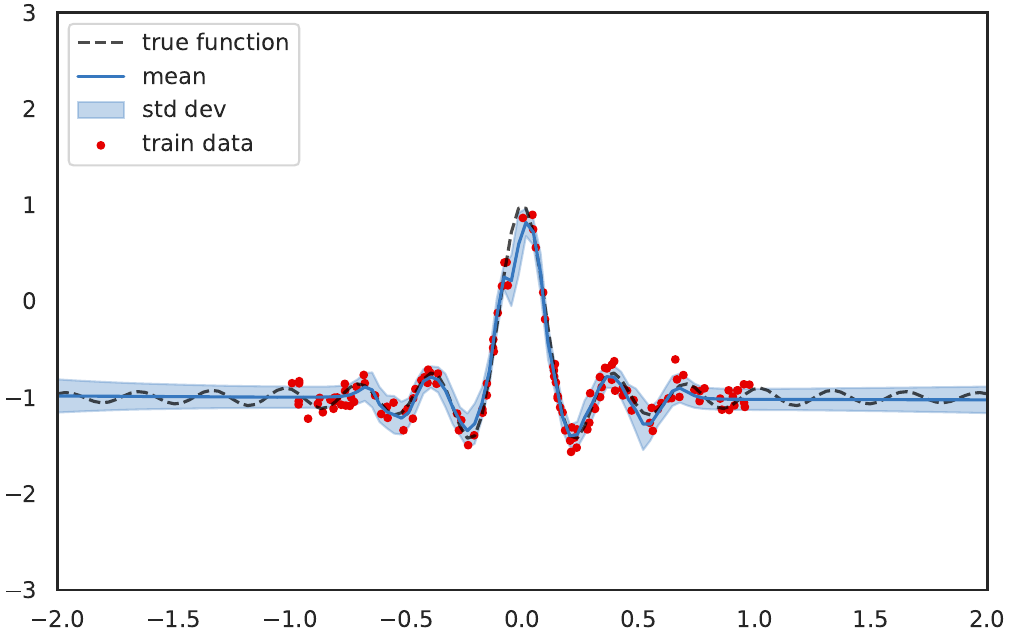}}
  \caption{Bayes By Backprop posterior-predictive distribution on the sinc dataset (cf.~Sec.~\ref{sec:experiments:toy}).}
  \label{fig:app:BBB_sinc}
\end{figure}

\begin{figure}[h!]
  \centering
  \subfigure[KL weighting factor $1.0$.]{\includegraphics[width=0.315\textwidth]{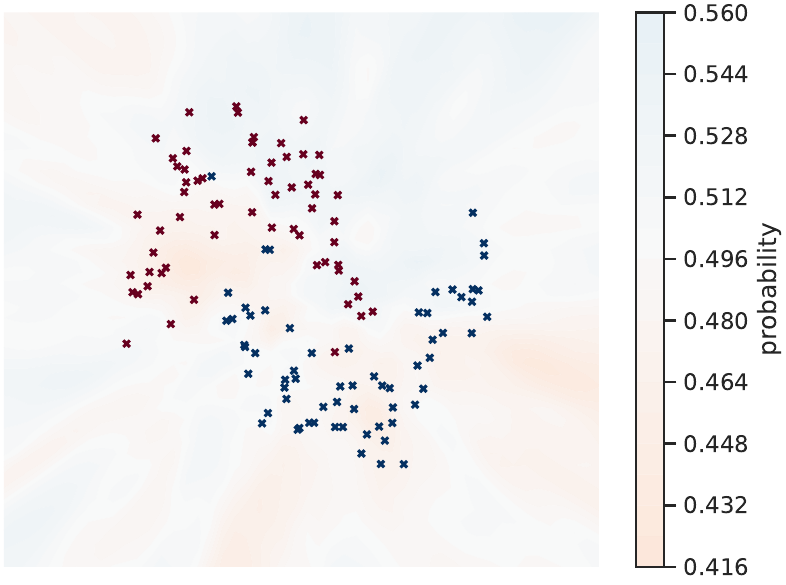}}
  \subfigure[KL weighting factor $0.1$.]{\includegraphics[width=0.315\textwidth]{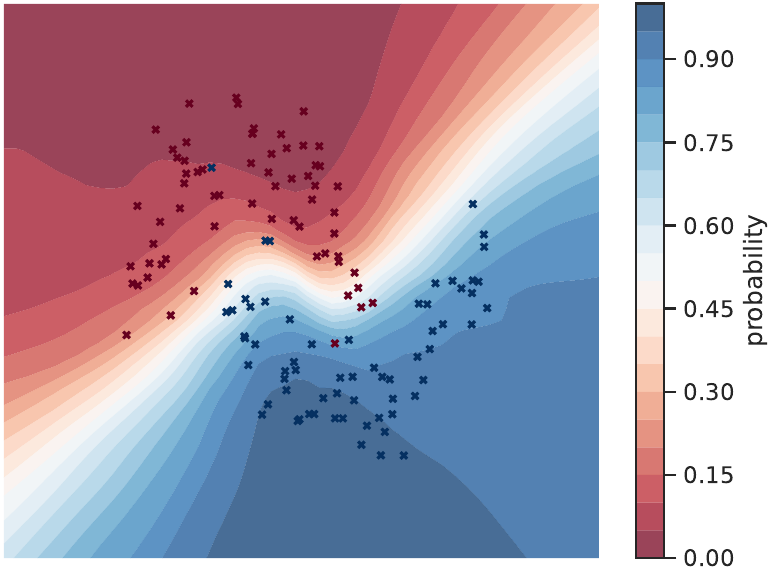}}
  \hfill
  \subfigure[KL weighting factor $0.01$.]{\includegraphics[width=0.315\textwidth]{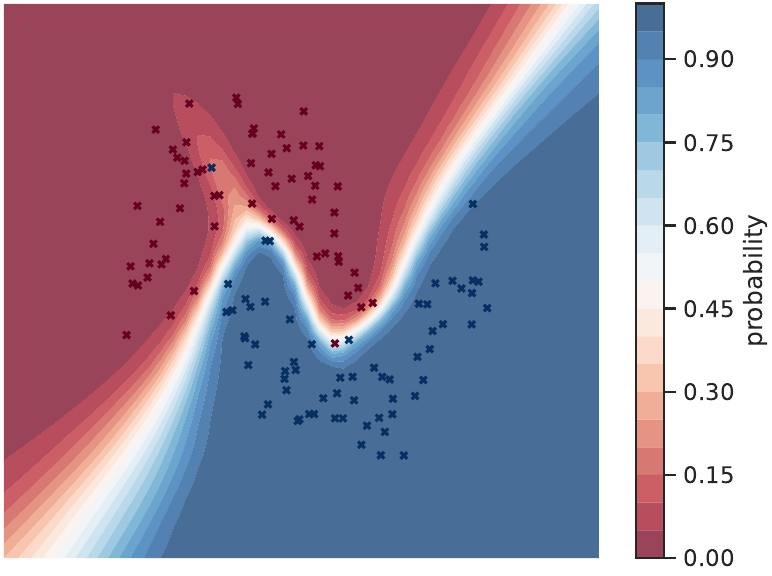}}
  \caption{Bayes By Backprop posterior-predictive distribution on the two-moons dataset (cf.~Sec.~\ref{sec:experiments:toy}).}
  \label{fig:app:BBB_moon}
\end{figure}

Here we replicate the experiments from Sec.~\ref{sec:experiments:toy}, using the same model architecture, however using standard variational inference (Bayes By Backprop). 
We plot the results for 3 different KL down-weighting factors in Fig.~\ref{fig:app:BBB_sine}, Fig.~\ref{fig:app:BBB_sinc} and  Fig.~\ref{fig:app:BBB_moon}.
As can be seen, inferring the weights using Bayes By Backprop results in both underfitting if the KL down-weighting factor is large and overconfidence if the factor is small.

\subsection{Local reparametrisation}
\begin{figure}[htb!]
  \centering
  \subfigure[FashionMNIST.]{
  	\includegraphics[width=0.48\linewidth]{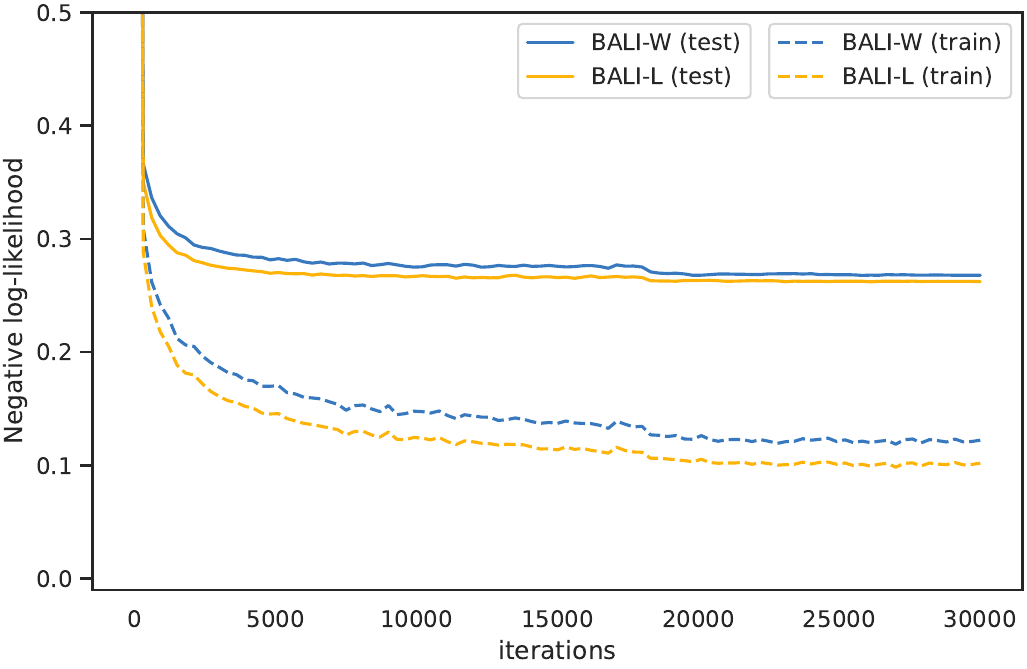}
  }
\hfill
  \subfigure[MNIST.]{
  	\includegraphics[width=0.48\linewidth]{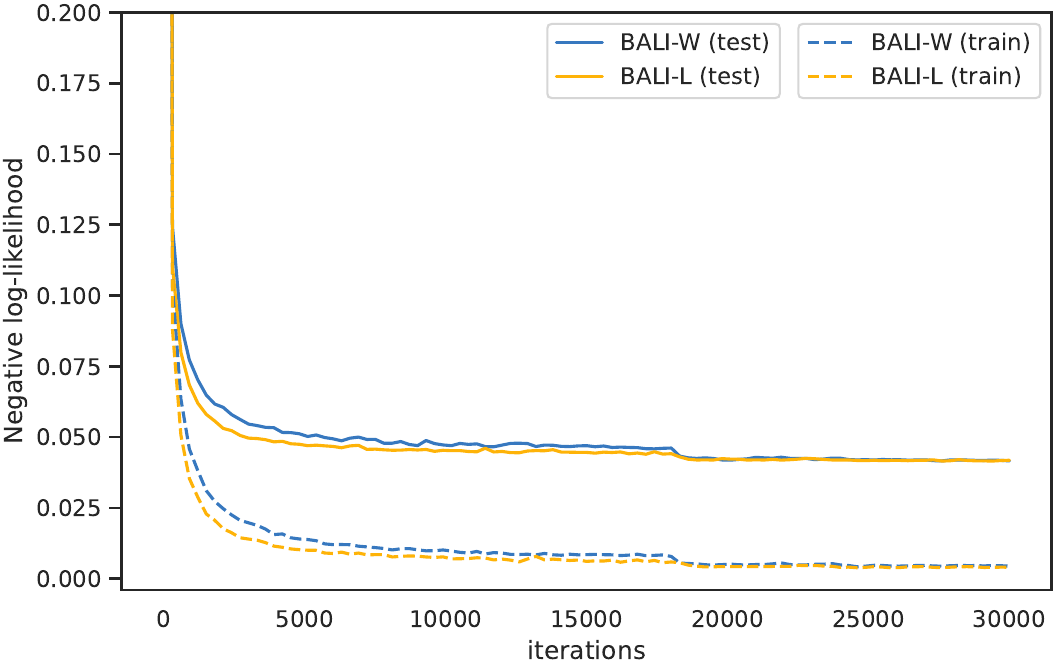}
  	}
\\
  \subfigure[Telescope.]{
  	\includegraphics[width=0.48\linewidth]{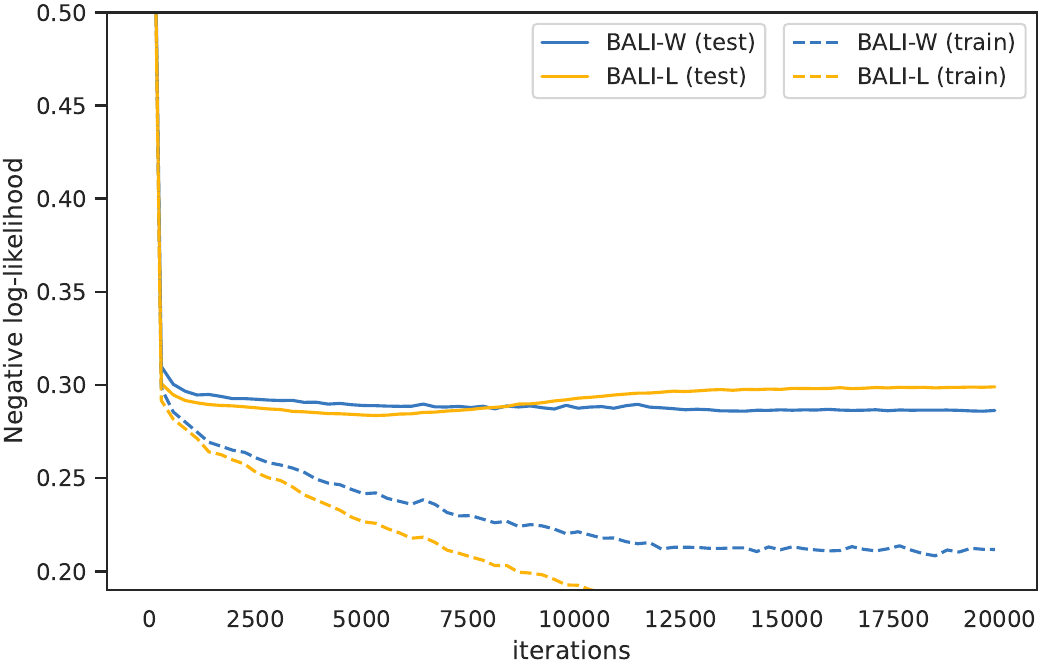}
  }
\hfill
  \subfigure[Spam.]{
  	\includegraphics[width=0.48\linewidth]{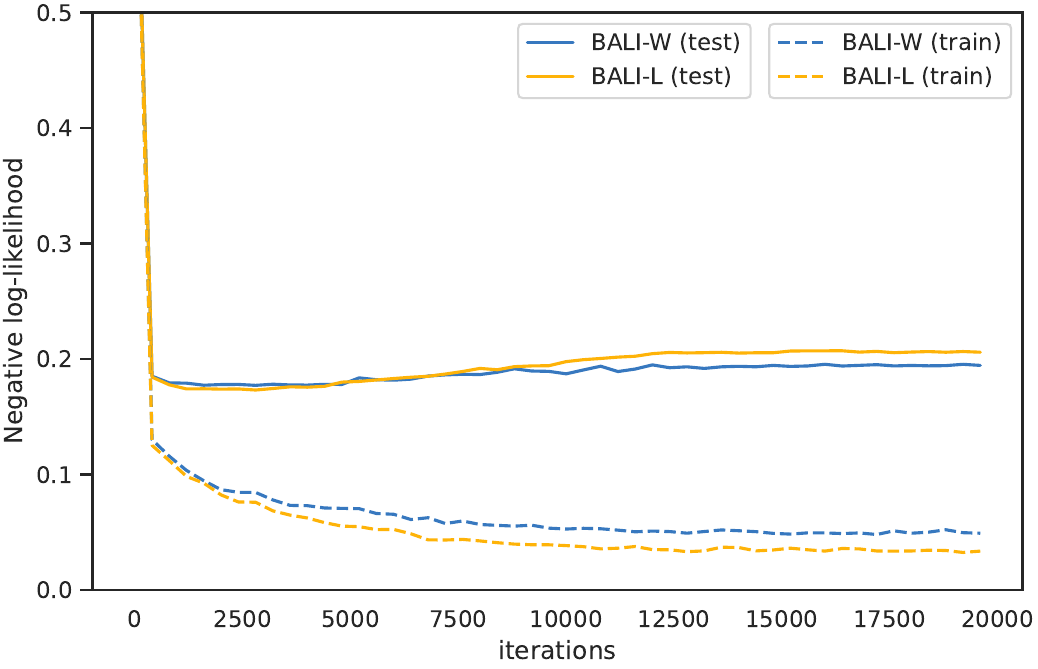}
  }
  \caption{
  Negative log-likelihood averaged over the test set (solid line) and training set (dashed line) over the course of training. 
  BALI-W refers to BALI used with weight reparametrisation as used in the main text and BALI-L refers to local reparametrisation as described in App.~\ref{app:noise_reparam}.
  }
  \label{fig:learning_curves_local}
\end{figure}

\makeatletter
\setlength{\@fptop}{0pt}
\makeatother
In App.~\ref{app:noise_reparam}, we have shown that we can define the pseudo-targets also via samples drawn from the posterior-predictive distribution in a given layer rather than sampling from the weights directly, which is similar to the local reparametrisation, extended to matrix-normal distributions. 
For the main paper, we have sampled from the weight distribution directly for simplicity. 
In Fig.~\ref{fig:learning_curves_local}, we show addtional results using local reparametrisation, referred to as BALI-L. 
As can be seen, local reparametrisation learns a bit faster, but also overfits stronger, potentially due to using less noise. 

%% file: bali.bbl
\begin{thebibliography}{10}

\bibitem{Ba2017Distributed}
Jimmy Ba, Roger Grosse, and James Martens.
\newblock Distributed second-order optimization using kronecker-factored
  approximations.
\newblock In {\em International Conference on Learning Representations}, 2017.

\bibitem{Ba16}
Lei~Jimmy Ba, Jamie~Ryan Kiros, and Geoffrey~E. Hinton.
\newblock Layer normalization.
\newblock {\em CoRR}, abs/1607.06450, 2016.

\bibitem{Benzing2022FOOF}
Frederik Benzing.
\newblock Gradient descent on neurons and its link to approximate second-order
  optimization.
\newblock In {\em International Conference on Machine Learning}, 2022.

\bibitem{Bickford2023prediction}
Freddie {Bickford Smith}, Andreas Kirsch, Sebastian Farquhar, Yarin Gal, Adam
  Foster, and Tom Rainforth.
\newblock Prediction-oriented {Bayesian} active learning.
\newblock {\em International Conference on Artificial Intelligence and
  Statistics}, 2023.

\bibitem{Blundell2015a}
Charles Blundell, Julien Cornebise, Koray Kavukcuoglu, and Daan Wierstra.
\newblock Weight uncertainty in neural networks.
\newblock In {\em Proceedings of the 32nd International Conference on
  International Conference on Machine Learning - Volume 37}, ICML'15, page
  1613–1622. JMLR.org, 2015.

\bibitem{Daxberger2021Laplace}
Erik Daxberger, Agustinus Kristiadi, Alexander Immer, Runa Eschenhagen,
  Matthias Bauer, and Philipp Hennig.
\newblock Laplace redux - effortless bayesian deep learning.
\newblock In M.~Ranzato, A.~Beygelzimer, Y.~Dauphin, P.S. Liang, and J.~Wortman
  Vaughan, editors, {\em Advances in Neural Information Processing Systems},
  volume~34, pages 20089--20103. Curran Associates, Inc., 2021.

\bibitem{Dusenberry2020Rank1Factors}
Michael~W. Dusenberry, Ghassen Jerfel, Yeming Wen, Yi-An Ma, Jasper Snoek,
  Katherine Heller, Balaji Lakshminarayanan, and Dustin Tran.
\newblock Efficient and scalable {B}ayesian neural nets with rank-1 factors.
\newblock In {\em Proceedings of the 37th International Conference on Machine
  Learning}, ICML'20. JMLR.org, 2020.

\bibitem{Farquhar2020Radial}
Sebastian Farquhar, Michael~A. Osborne, and Yarin Gal.
\newblock Radial bayesian neural networks: Beyond discrete support in
  large-scale bayesian deep learning.
\newblock In Silvia Chiappa and Roberto Calandra, editors, {\em Proceedings of
  the Twenty Third International Conference on Artificial Intelligence and
  Statistics}, volume 108 of {\em Proceedings of Machine Learning Research},
  pages 1352--1362. PMLR, 26--28 Aug 2020.

\bibitem{Gal2016}
Yarin Gal.
\newblock {\em Uncertainty in Deep Learning}.
\newblock PhD thesis, University of Cambridge, 2016.

\bibitem{Gal2017Active}
Yarin Gal, Riashat Islam, and Zoubin Ghahramani.
\newblock Deep bayesian active learning with image data.
\newblock In {\em Proceedings of the 34th International Conference on Machine
  Learning - Volume 70}, ICML'17, page 1183–1192. JMLR.org, 2017.

\bibitem{Ghosh2018}
Soumya Ghosh, Jiayu Yao, and Finale Doshi-Velez.
\newblock Structured variational learning of {B}ayesian neural networks with
  horseshoe priors.
\newblock In Jennifer Dy and Andreas Krause, editors, {\em Proceedings of the
  35th International Conference on Machine Learning}, volume~80 of {\em
  Proceedings of Machine Learning Research}, pages 1744--1753,
  Stockholmsmässan, Stockholm Sweden, 2018. PMLR.

\bibitem{Graves2011}
Alex Graves.
\newblock Practical variational inference for neural networks.
\newblock In J.~Shawe-Taylor, R.~Zemel, P.~Bartlett, F.~Pereira, and K.~Q.
  Weinberger, editors, {\em Advances in Neural Information Processing Systems},
  volume~24, pages 2348--2356. Curran Associates, Inc., 2011.

\bibitem{Hernandez2015PBP}
Jose~Miguel Hernandez-Lobato and Ryan Adams.
\newblock Probabilistic backpropagation for scalable learning of bayesian
  neural networks.
\newblock In Francis Bach and David Blei, editors, {\em Proceedings of the 32nd
  International Conference on Machine Learning}, volume~37 of {\em Proceedings
  of Machine Learning Research}, pages 1861--1869, Lille, France, 07--09 Jul
  2015. PMLR.

\bibitem{Hinton1993a}
Geoffrey~E. Hinton and Drew van Camp.
\newblock Keeping the neural networks simple by minimizing the description
  length of the weights.
\newblock In {\em Proceedings of the Sixth Annual Conference on Computational
  Learning Theory}, COLT '93, page 5–13. Association for Computing Machinery,
  1993.

\bibitem{Ioffe2015}
Sergey Ioffe and Christian Szegedy.
\newblock Batch normalization: Accelerating deep network training by reducing
  internal covariate shift.
\newblock ICML'15, page 448–456. JMLR.org, 2015.

\bibitem{khan2018fast}
Mohammad~Emtiyaz Khan, Didrik Nielsen, Voot Tangkaratt, Wu~Lin, Yarin Gal, and
  Akash Srivastava.
\newblock Fast and scalable bayesian deep learning by weight-perturbation in
  adam, 2018.

\bibitem{Kingma2014Adam}
Diederik~P. Kingma and Jimmy Ba.
\newblock Adam: {A} method for stochastic optimization.
\newblock In Yoshua Bengio and Yann LeCun, editors, {\em 3rd International
  Conference on Learning Representations, {ICLR} 2015, San Diego, CA, USA, May
  7-9, 2015, Conference Track Proceedings}, 2015.

\bibitem{Kingma2015LocalReparam}
Diederik~P. Kingma, Tim Salimans, and Max Welling.
\newblock Variational dropout and the local reparameterization trick.
\newblock In {\em Proceedings of the 28th International Conference on Neural
  Information Processing Systems - Volume 2}, NIPS'15, page 2575–2583. MIT
  Press, 2015.

\bibitem{Kingma2013VAE}
Diederik~P. Kingma and Max Welling.
\newblock Auto-encoding variational {B}ayes.
\newblock In Yoshua Bengio and Yann LeCun, editors, {\em 2nd International
  Conference on Learning Representations, {ICLR} 2014, Banff, AB, Canada, April
  14-16, 2014, Conference Track Proceedings}, 2014.

\bibitem{Kurle2020LLL}
Richard Kurle, Botond Cseke, Alexej Klushyn, Patrick van~der Smagt, and Stephan
  G{\"{u}}nnemann.
\newblock Continual learning with {B}ayesian neural networks for non-stationary
  data.
\newblock In {\em 8th International Conference on Learning Representations,
  {ICLR} 2020, Addis Ababa, Ethiopia, April 26-30, 2020}. OpenReview.net, 2020.

\bibitem{Kurle2022VBNN}
Richard Kurle, Ralf Herbrich, Tim Januschowski, Yuyang~(Bernie) Wang, and Jan
  Gasthaus.
\newblock On the detrimental effect of invariances in the likelihood for
  variational inference.
\newblock In S.~Koyejo, S.~Mohamed, A.~Agarwal, D.~Belgrave, K.~Cho, and A.~Oh,
  editors, {\em Advances in Neural Information Processing Systems}, volume~35,
  pages 4531--4542. Curran Associates, Inc., 2022.

\bibitem{Kurle2021BDL}
Richard Kurle, Tim Januschowski, Jan Gasthaus, and Bernie Wang.
\newblock On symmetries in variational {B}ayesian neural nets.
\newblock In {\em {B}ayesian Deep Learning NeurIPS workshop}, 2021.

\bibitem{Loan2000Kron}
Charles~F.Van Loan.
\newblock The ubiquitous kronecker product.
\newblock {\em Journal of Computational and Applied Mathematics},
  123(1):85--100, 2000.
\newblock Numerical Analysis 2000. Vol. III: Linear Algebra.

\bibitem{MacKay1992a}
David J.~C. MacKay.
\newblock A practical {B}ayesian framework for backpropagation networks.
\newblock {\em Neural Comput.}, 4(3):448–472, 1992.

\bibitem{Martens2020NG}
James Martens.
\newblock New insights and perspectives on the natural gradient method.
\newblock {\em J. Mach. Learn. Res.}, 21(1), jan 2020.

\bibitem{Martens2015KFAC}
James Martens and Roger Grosse.
\newblock Optimizing neural networks with kronecker-factored approximate
  curvature.
\newblock In {\em Proceedings of the 32nd International Conference on
  International Conference on Machine Learning - Volume 37}, ICML'15, page
  2408–2417. JMLR.org, 2015.

\bibitem{MurphyProbabilisticPerspective}
Kevin~P. Murphy.
\newblock {\em Machine Learning: A Probabilistic Perspective}.
\newblock The MIT Press, 2012.

\bibitem{Nguyen2018VCL}
Cuong~V. Nguyen, Yingzhen Li, Thang~D. Bui, and Richard~E. Turner.
\newblock Variational continual learning.
\newblock In {\em International Conference on Learning Representations}, 2018.

\bibitem{Ober2021GlobalInducingPoint}
Sebastian~W Ober and Laurence Aitchison.
\newblock Global inducing point variational posteriors for bayesian neural
  networks and deep gaussian processes.
\newblock In Marina Meila and Tong Zhang, editors, {\em Proceedings of the 38th
  International Conference on Machine Learning}, volume 139 of {\em Proceedings
  of Machine Learning Research}, pages 8248--8259. PMLR, 18--24 Jul 2021.

\bibitem{Osawa2019}
Kazuki Osawa, Siddharth Swaroop, Mohammad Emtiyaz~E Khan, Anirudh Jain, Runa
  Eschenhagen, Richard~E Turner, and Rio Yokota.
\newblock Practical deep learning with {B}ayesian principles.
\newblock In H.~Wallach, H.~Larochelle, A.~Beygelzimer, F.~d\textquotesingle
  Alch\'{e}-Buc, E.~Fox, and R.~Garnett, editors, {\em Advances in Neural
  Information Processing Systems}, volume~32. Curran Associates, Inc., 2019.

\bibitem{Ritter2018LA}
Hippolyt Ritter, Aleksandar Botev, and David Barber.
\newblock Online structured laplace approximations for overcoming catastrophic
  forgetting.
\newblock In {\em Proceedings of the 32nd International Conference on Neural
  Information Processing Systems}, NeurIPS 2018, page 3742–3752, Red Hook,
  NY, USA, 2018. Curran Associates Inc.

\bibitem{Shi2021Distributed}
Shaohuai Shi, Lin Zhang, and Bo~Li.
\newblock Accelerating distributed k-fac with smart parallelism of computing
  and communication tasks.
\newblock In {\em 2021 IEEE 41st International Conference on Distributed
  Computing Systems (ICDCS)}, pages 550--560, 2021.

\bibitem{Swiatkowski2020}
Jakub Swiatkowski, Kevin Roth, Bastiaan~S. Veeling, Linh Tran, Joshua~V.
  Dillon, Jasper Snoek, Stephan Mandt, Tim Salimans, Rodolphe Jenatton, and
  Sebastian Nowozin.
\newblock The k-tied normal distribution: A compact parameterization of
  gaussian mean field posteriors in {B}ayesian neural networks.
\newblock In {\em Proceedings of the 37th International Conference on Machine
  Learning}, ICML'20. JMLR.org, 2020.

\bibitem{Tomczak2020LowRank}
Marcin Tomczak, Siddharth Swaroop, and Richard Turner.
\newblock Efficient low rank gaussian variational inference for neural
  networks.
\newblock In H.~Larochelle, M.~Ranzato, R.~Hadsell, M.F. Balcan, and H.~Lin,
  editors, {\em Advances in Neural Information Processing Systems}, volume~33,
  pages 4610--4622. Curran Associates, Inc., 2020.

\bibitem{Tomczak2021Collapsed}
Marcin~B. Tomczak, Siddharth Swaroop, Andrew Y.~K. Foong, and Richard~E Turner.
\newblock Collapsed variational bounds for {B}ayesian neural networks.
\newblock In A.~Beygelzimer, Y.~Dauphin, P.~Liang, and J.~Wortman Vaughan,
  editors, {\em Advances in Neural Information Processing Systems}, 2021.

\bibitem{Trippe2017}
Brian Trippe and Richard Turner.
\newblock Overpruning in variational {B}ayesian neural networks.
\newblock {\em arXiv preprint arXiv:1801.06230}, 2018.

\bibitem{Wilson2022Bayesian}
Andrew~G Wilson and Pavel Izmailov.
\newblock Bayesian deep learning and a probabilistic perspective of
  generalization.
\newblock In H.~Larochelle, M.~Ranzato, R.~Hadsell, M.F. Balcan, and H.~Lin,
  editors, {\em Advances in Neural Information Processing Systems}, volume~33,
  pages 4697--4708. Curran Associates, Inc., 2020.

\bibitem{Zhang18NoisyKFAC}
Guodong Zhang, Shengyang Sun, David Duvenaud, and Roger Grosse.
\newblock Noisy natural gradient as variational inference.
\newblock In Jennifer Dy and Andreas Krause, editors, {\em Proceedings of the
  35th International Conference on Machine Learning}, volume~80 of {\em
  Proceedings of Machine Learning Research}, pages 5852--5861. PMLR, 10--15 Jul
  2018.

\bibitem{Zhang2013Kronecker}
Huamin Zhang and Feng Ding.
\newblock {On the Kronecker Products and Their Applications}.
\newblock {\em Journal of Applied Mathematics}, 2013(none):1 -- 8, 2013.

\bibitem{Zhang2022Scalable}
Lin Zhang, Shaohuai Shi, Wei Wang, and Bo~Li.
\newblock Scalable k-fac training for deep neural networks with distributed
  preconditioning, 2022.

\end{thebibliography}
